\documentclass[10pt,twocolumn,letterpaper]{article}

\usepackage{iccv}
\usepackage{times}
\usepackage{epsfig}
\usepackage{graphicx}
\usepackage{amsmath}
\usepackage[dvipsnames]{xcolor}
\usepackage{amssymb}
\usepackage{booktabs}
\usepackage{multicol}
\usepackage{multirow}
\usepackage{tabu}
\usepackage[super]{nth}
\usepackage{enumitem}
\usepackage{caption}
\usepackage{pifont}
\usepackage{algorithm}
\usepackage{algpseudocode}

\usepackage[dvipsnames]{xcolor}

\definecolor{mygreen}{RGB}{126, 180, 78}

\newcommand{\Yunze}[1]{\textcolor{red}{[Yunze: #1]}}

\newcommand{\intensity}{reflectance\xspace}
\newcommand{\Intensity}{Reflectance\xspace}

\newcommand{\ambient}{\mathsf{ambient}}
\newcommand{\reflectance}{\mathsf{reflectance}}

\DeclareMathAlphabet{\mathpzc}{T1}{pzc}{m}{it}

\usepackage[pagebackref=true,breaklinks=true,letterpaper=true,colorlinks,bookmarks=false]{hyperref}

\iccvfinalcopy 

\ificcvfinal\fi

\begin{document}

\title{Multi-Echo LiDAR for 3D Object Detection}

\author{Yunze Man\textsuperscript{1}, Xinshuo Weng\textsuperscript{1}, Prasanna Kumar Sivakumar\textsuperscript{2}, Matthew O'Toole\textsuperscript{1}, Kris Kitani\textsuperscript{1}\\
{\textsuperscript{1}Carnegie Mellon University, \textsuperscript{2}DENSO} \\ 
{\tt\small \{yman, xinshuow, mpotoole, kkitani\}@cs.cmu.edu, \ \ prasannakumar.0091@gmail.com}
}

\maketitle
\thispagestyle{empty}


\begin{abstract}
LiDAR sensors can be used to obtain a wide range of measurement signals other than a simple 3D point cloud, and those signals can be leveraged to improve perception tasks like 3D object detection. A single laser pulse can be partially reflected by multiple objects along its path, resulting in multiple measurements called echoes. Multi-echo measurement can provide information about object contours and semi-transparent surfaces which can be used to better identify and locate objects. LiDAR can also measure surface reflectance (intensity of laser pulse return), as well as ambient light of the scene (sunlight reflected by objects). These signals are already available in commercial LiDAR devices but have not been used in most LiDAR-based detection models.
We present a 3D object detection model which leverages the full spectrum of measurement signals provided by LiDAR. First, we propose a multi-signal fusion (MSF) module to combine (1) the reflectance and ambient features extracted with a 2D CNN, and (2) point cloud features extracted using a 3D graph neural network (GNN). Second, we propose a multi-echo aggregation (MEA) module to combine the information encoded in different set of echo points. Compared with traditional single echo point cloud methods, our proposed Multi-Signal LiDAR Detector (MSLiD) extracts richer context information from a wider range of sensing measurements and achieves more accurate 3D object detection. Experiments show that by incorporating the multi-modality of LiDAR, our method outperforms the state-of-the-art by up to 9.1\%.  
\end{abstract}

\begin{figure}[!t]
\begin{center}
\includegraphics[trim=455 203 260 222,clip, width=\linewidth]{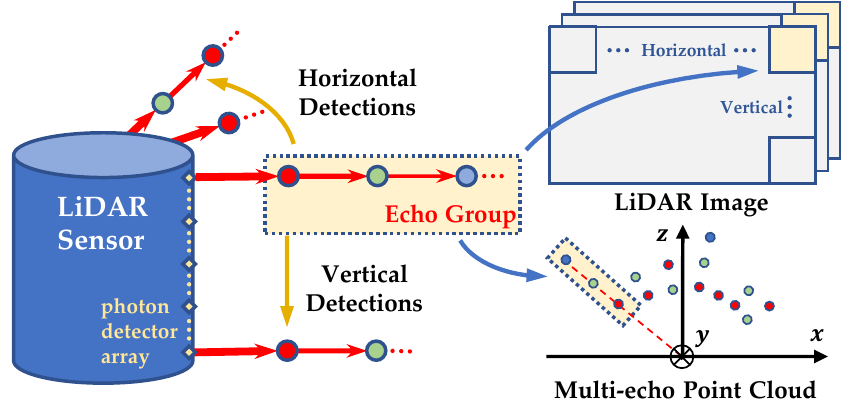}
\end{center}
\caption{\textbf{Illustration of multi-signal measurements from LiDAR sensor.} Each photon detector on the sensor collects a group of signals and forms an "Echo Group", which is converted into a 2D LiDAR image and a multi-echo point cloud representation. } 
\label{fig:teaser}
\vspace{-0.2cm}
\end{figure}

\section{Introduction}

LiDAR is a powerful sensor which has the ability to capture a wide range measurements for perception tasks including object detection. The most commonly used LiDAR measurement type is a set of 3D points (a point cloud) and their \intensity values, which provides accurate 3D shape information of objects in the scene. State-of-the-art object detection methods have made great breakthroughs by leveraging 3D point cloud data. However, despite such success, there are several types of LiDAR measurements which are largely ignored in modern day LiDAR perception algorithms. In the following, we describe three unique features of the LiDAR sensor, which are available in standard LiDAR sensors, but surprisingly have rarely been used in published LiDAR-based object detection algorithms. We show that by leveraging these features, one can greatly improve 3D object detection performance.


The first important feature of LiDAR is its ability to obtain multiple return signals with a single laser pulse, called echoes. LiDAR is a time-of-flight measurement system which measures the time it takes for a laser pulse to hit an object and return to the sensor. More specifically, the laser emits a short pulse, and a photodetector timestamps the arrival of photons reflected back by object surfaces. It is possible for a photodetector to acquire multiple return signals (echoes) if the laser is partially reflected by multiple objects along its path of propagation. We call the multiple returned signals generated from the same laser beam an `echo group.' Points in the same echo group lie on one line in 3D space, and they are typically ordered according to their signal strength. In addition to the direct benefit of increasing the number of points available, multiple echoes also imply that the laser signal is only partially reflected at non-terminal echo locations. Thus, these echo points are likely on the contour of an object (it obstructs only a part of the laser) or on a semi-transparent surface (a portion of the laser propagates through the surface). In either case, we hypothesize that echoes encode meaningful features that can help locate or classify an object.

The second important feature of LiDAR is the ability to capture ambient scene illumination. The photodetector of the LiDAR continuously captures infrared (IR) light and therefore is capturing IR images of the scene (typically reflected sunlight) between laser pulses. Although this information is typically ignored in most LiDAR based perception algorithms, a LiDAR can be used to capture an image of the scene using the IR spectrum. Ambient measurements can be processed as a 2D image and can be used to extract texture information about the objects in the scene.

The third important features of LiDAR is the ability to capture surface \intensity. LiDAR captures laser signal returns, so each point will have a corresponding \intensity value which measures the strength of the detected laser pulse. \Intensity also encodes material properties of objects useful for detection and classification. Unlike the ambient signal, different points inside the same echo group will have different \intensity values, resulting in multiple \intensity intensity values which we call multi-echo reflectance.

We propose a multi-signal LiDAR-based 3D object detector (MSLiD). First, in order to better leverage the dense texture and surface properties encoded in the ambient and reflectance signals, we re-organize them as a dense 2D representation, called the `LiDAR Image.' Then, in order to combine the dense 2D `LiDAR Image' with sparse 3D point cloud, we propose a multi-signal fusion (MSF) module which incorporates a 2D CNN branch and a 3D GNN branch. The MSF module aims at fusing 2D visual information with 3D positional information by sending pixel and class feature from 2D branch to point-wise feature learned in 3D branch. Furthermore, In order to extract and combine information encoded in different echo groups, 
we propose a multi-echo aggregation (MEA) module. To resolve the imbalance between number of points in different echos, The MEA module reassigns multi-echo points into two sets -- `penetrable' and `impenetrable' sets according to whether the object reflects partial laser signal. Aggregating the features learned from two new sets of points provides a richer context information of objects and leads to better location estimation. 
By cascading the MSF and MEA modules, the proposed system combines dense visual information from ambient/reflectance and sparse geometric information from point cloud, while also extract richer context feature by aggregating multiple echoes. By leveraging multi-signal LiDAR measurements other than a single point cloud, MSLiD learns a more discriminative object representation which leads to accurate object localization and classification.

We collect one real-world and one synthetic dataset with multiple LiDAR measurements, including ambient signal, multi-echo point cloud and reflectance signals. Experiments on two datasets demonstrate that our method outperforms state-of-the-art single-echo methods by up to $9.1$\%. Overall, our contributions can be summarized as follows:
\begin{enumerate}
    \item MSLiD is the first to propose a 3D detection framework that properly leverages ambient illumination, multiple echos of point clouds and reflectance signals for LiDAR sensor. Our method shows improvement over prior methods using single-echo point cloud with reflectance intensity.
    \item We propose a multi-signal fusion module to effectively combine dense visual information from ambient and reflectance signals with sparse 3D positional information from point cloud.
    \item We propose a multi-echo aggregation module to form a richer context representation of objects from multiple groups of echoes, resulting in more accurate object localization and classification. 
\end{enumerate}


\begin{figure*}[!t]
\begin{center}
\includegraphics[trim= 58 128 68 122 , clip=true, width=\linewidth]{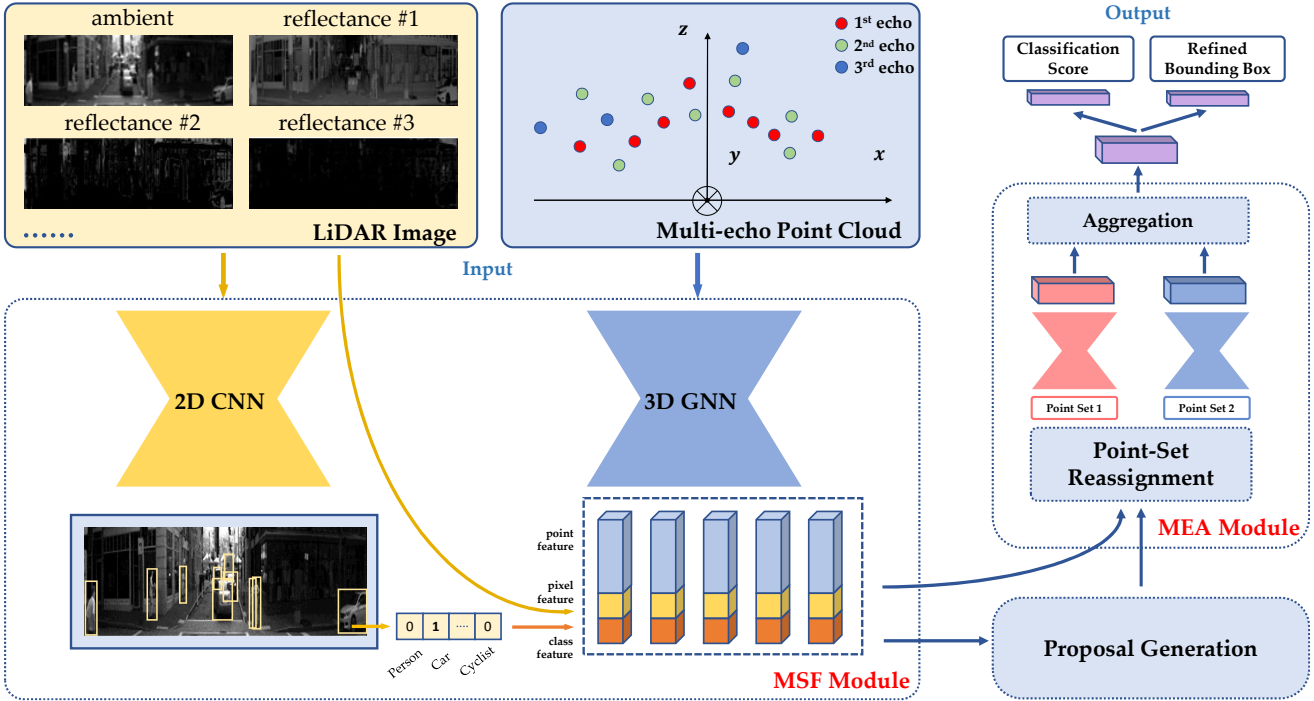}
\end{center}
\caption{Illustration of our proposed framework. Our method takes multi-echo point cloud and LiDAR image as input. The MSF module learns separate features for 2D and 3D representation, and fuses 2D pixel feature with 3D point feature. The MEA module first performs point set reassignment to group together points with similar properties. Discriminative RoI feature for each proposal candidate is then learned by aggregating features learned from different set of points. The learned feature is then used for confidence estimation and bounding box regression. } 
\label{fig:pipeline}
\vspace{-0.2cm}
\end{figure*}

\section{Related Work} 

\noindent\textbf{3D Object Detection with Grid-based Methods.} 
Many existing works convert the point cloud into a regular grid space representation in order to tackle the inherent sparsity and irregular format of the point cloud. \cite{cmw17mv3d,kml18avod,lyw18continuous,ylu18hdnet} project 3D point cloud into 2D birds eye view image to extract feature with mature 2D CNN. For real-time detection, \cite{ylu18pixor,lvc19pointpillars} explore more efficient framework for birds eye view transformation. Other work focuses on 3D voxel representation. \cite{zto18voxelnet} voxelizes the point cloud and uses 3D CNNs to extract features. Sparse convolution \cite{gev18sparse} is introduced in \cite{yml18second} for more efficient voxel processing and feature extraction. Also, \cite{wj19frustum} explores non-regular shape of 3D voxels while \cite{sgj20pvrcnn} proposes to combine point feature learning with voxelization, leading to higher detection performance. Grid-based methods are generally efficient for proposal generation, but they suffer from information loss during the projection or voxelization process. 
In contrast, our method does not have the problem of loss of information as we do not voxelize or project point clouds.

\vspace{1.5mm}\noindent\textbf{3D Object Detection with Point-based Methods.} 
F-PointNet \cite{qlw18fpointnet} first proposes to use frustum proposals from 2D object detection and regresses the final bounding box directly from point features extracted by PointNet \cite{qsm17pointnet, qys17pointnet++}. \cite{swl19pointrcnn} instead proposes to generate 3D candidate proposals directly from the point cloud in a bottom-up manner, and the following \cite{ysl19std} proposes to learn a dense voxel representation of each candidate proposal for more efficient bounding box regression. \cite{ysl203dssd} further reduces the inference time by removing the refinement stage and regressing directly from 3D keypoints. Despite using different point cloud encoding frameworks, these works are working with single-echo point cloud and \intensity value. Compared to prior work, our method leverages multi-echo point clouds and ambient images to learn a richer representation for both point-level features and proposal-level features.

\vspace{1.5mm}\noindent\textbf{3D Object Detection with Multi-modal Fusion Methods.} 
Exploring effective ways to fuse signals from multiple modalities is still an open question in 3D object detection. \cite{kml18avod, cmw17mv3d, lyc19mtms} propose to project point cloud to BEV space, and then fuse 2D RGB features with a BEV feature to generate proposals and regress bounding boxes. \cite{qlw18fpointnet} does not apply feature fusion, but instead uses 2D detection bounding boxes to guide 3D proposal generation. \cite{zmp20cross, ykk203dcvf, hlc20epnet} explores deep feature fusion between LiDAR and RGB sensors. \cite{vlh20pointpainting} proposes to augment point cloud with image semantic segmentation results. \cite{qcl20imvotenet} proposes to fuse geometry cue, semantic cue and texture cue from 2D features with 3D point features and obtains promising performance for indoor 3D object detection. \cite{sdh20clocs} proposed a post-detection fusion mechanism to combine the candidate boxes from RGB and LiDAR inputs. In our work, we propose to work with a wider range of modalities coming from the LiDAR sensor. We are the first to explore a proper way of combining multi-echo points with ambient/\intensity signal information.  
 
\section{Approach}

We design a 3D object detection solution suited for multi-signal measurements provided by the LiDAR sensor, including multi-echo point cloud, as well as ambient and \intensity signals. Our Multi-Signal LiDAR-based Detector, called MSLiD, achieves fusion between multiple LiDAR signals with two modules, \textit{i.e.} Multi-Signal Fusion (MSF) module where visual signals are combined with geometric signals, and Multi-Echo Aggregation (MEA) module where points from different echo groups are fused together. In this section, we first describe how multiple types of signals are processed, encoded and fused in the MSF module for point-wise feature extraction and proposal generation (Sec~\ref{sec:MSF}). Then, we describe the MEA module where multiple point clouds are reassigned into different point sets, and features of these point sets are extracted and aggregated to form the proper proposal RoI feature for 3D bounding box regression (Sec~\ref{sec:mea}). Note that we assume the LiDAR sensor generates $k$-echo point clouds. The overall pipeline of our method is shown in Fig~\ref{fig:pipeline}.

\subsection{Multi-Signal Fusion for Proposal Generation} \label{sec:MSF}

Learning features of multiple modalities by multiple streams \cite{lyc19mtms, qcl20imvotenet, vlh20pointpainting} is proved to be effective in feature fusion. Aiming at fully making use of the complementary nature between different signals, we utilize a two-stream feature extraction and blending framework.

\vspace{1.5mm}
\noindent \textbf{Data Encoding.} 
In order to leverage mature 2D detection models to extract visual cues from multiple signals, we first convert ambient and reflectance signals into 2D image representation called "LiDAR image". Unlike previous methods where point cloud is projected into image space with the calibration matrix, we re-organize the LiDAR multi-modal measurement into a 2D image according to the alignment of LiDAR detector array (\emph{i.e.}, range view \cite{Caccia2019}), as shown in Fig~\ref{fig:teaser}. Specifically, each column of pixels are signals captured by the vertically aligned LiDAR detectors and each row of pixels are signals captured by the same detector in different horizontal directions. The converted LiDAR image has resolution $[h, w, n]$ where $h$ corresponds to the number of vertically aligned strip of detectors and $w$ corresponds to horizontally aligned ones. $n$ is the number of modalities encoded in each `pixel', which in our case includes one ambient value and $k$ \intensity values. 

For the image stream, the converted LiDAR images are passed into a 2D detector to generate 2D bounding boxes. We adopt a FPN-based model \cite{lpg17fpn} as our 2D detector, where the backbone weights are pre-trained on ImageNet classification and the model is pre-trained on COCO object detection. We then fine-tune it on our dataset using LiDAR image as inputs to detect 2D bounding boxes. For the point cloud stream, we utilize PointNet++ \cite{qys17pointnet++} as our backbone network to learn discriminative point-wise features from raw 3D point cloud. To leverage point clouds from different echos and increase point density, we group $k$-echo points together as a whole point cloud and extract features from it. 

\vspace{1.5mm}
\noindent \textbf{Multi-Signal Fusion (MSF).} 
In order to generate rich point-wise features for proposal generation, we present a 2D-to-3D feature blending method that augments 2D pixel-wise semantic information to 3D point-wise geometric features. \cite{qcl20imvotenet} presents an effective indoor 3D object detection framework to extract three different cues (features) from 2D detection, where geometric cue infers 3D proposal center from 2D bounding box center, semantics cue infers the object type of the proposal, and texture cue (RGB value) encodes the texture information of the surface. However, the geometric cue relies on a very strict assumption where the sensor origin, 2D center and 3D center lies on the same line. This assumption approximately holds for indoor scenarios with controllable depth range and object shape. But for the autonomous driving scenario, the error caused by this approximation is often unbearable.   

In contrast, we propose an MSF module which appends 2D semantic features from a pixel to its corresponding point - class probability vector (class vector) and dense LiDAR measurements (pixel vector). Specifically, for each pixel of the LiDAR image, we form a one-hot vector to represent which class it belongs to. If the pixel is not inside any bounding box, the vector is set to be all zero, and if the pixel is inside multiple bounding boxes, the corresponding class entries are all set to one. We believe that the regional prediction vector helps to address the ambiguity of object class in the sparse 3D point cloud. On the other hand, the dense LiDAR measurements include the ambient value and \intensity value the point corresponds to. Each point has its unique \intensity value, while multiple points in the same echo group share the same ambient value. The dense LiDAR measurements include lower-level semantic features that 3D point cloud does not possess, including object texture and surface reflectivity. By fusing 2D semantic features with 3D point features, our method can encode multi-modal information which helps better locate and identify objects.  

Given fused features, we apply a bottom-up proposal generation strategy \cite{swl19pointrcnn} to generate 3D candidates for the bounding box refinement stage. Specifically, we learn a binary foreground/background segmentation using the fused point-wise features, where foreground points are defined as points inside any ground truth 3D bounding boxes. Afterward, we generate anchor boxes around foreground points and regress the residual of box parameters \cite{swl19pointrcnn, qlw18fpointnet}.

\subsection{Multi-Echo Aggregation for Box Refinement}\label{sec:mea}

Given several proposal candidates, the bounding box refinement network aims to estimate the proposal confidence and predict the residuals of bounding box parameters (i.e., center, size and orientation) based on a representative RoI feature. Following \cite{swl19pointrcnn, ysl19std, sws20parta2}, we first perform a canonical transformation to each point by subtracting their 3D position with the proposal center $(X, Y, Z)$ values and rotating them to the proposal predicted orientation. This makes the model robust under geometrical transformation and thus can learn better local features. Then we want to learn a representative RoI feature of each proposal for confidence estimation and bounding box regression. Our main motivation is to effectively extract discriminative features from multi-echo point clouds. To this end, we propose point set reassignment and multi-echo aggregation modules.

\vspace{1.5mm}
\noindent \textbf{Point Set Reassignment.} \label{sec:reassign}
The most straightforward way to leverage multi-echo points in RoI feature pooling is to learn separate features for each group of echo points. However, this naive idea leads to poor performance for two reasons. The first reason is that the numbers of points in different echo point clouds are extremely biased. If the echoes are ordered by signal \intensity (where low order echoes have higher intensity), then higher echo point clouds will have far fewer points than low echo point clouds. This is because points with lower signal intensity are less likely to be detected by the sensor. Second, too many uncontrollable factors can affect the order of echo points. Since echoes are ordered according to signal \intensity, various unpredictable factors decide with echo group a point belongs to, including but not limited to atmosphere humidity, incident angle, surface roughness and other optical properties of the object surface. It turns out the network is not able to extract useful features to refine the bounding box when a huge pack of factors is entangled in every each echo point cloud.

\begin{figure}[!t]
\begin{center}
\includegraphics[width=\linewidth]{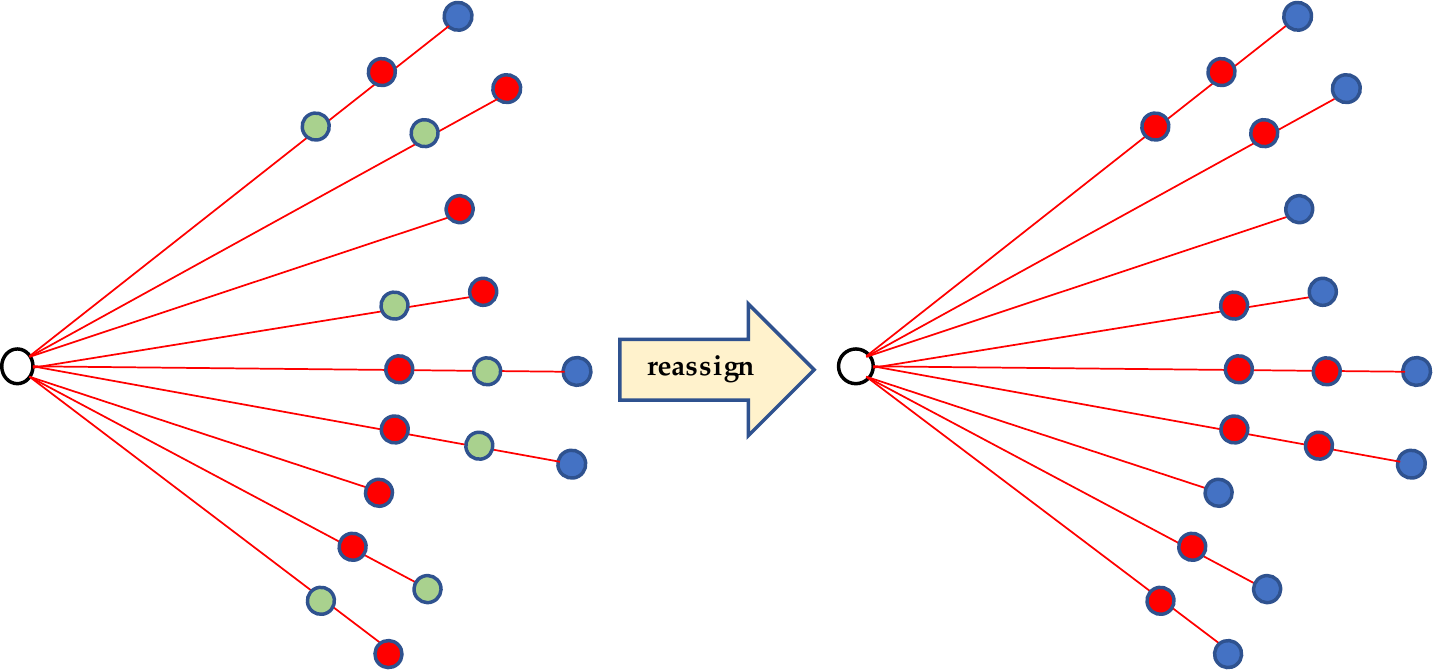}
\end{center}
\caption{Illustration of our point reassignment strategy on a three-echo point cloud, where white dots represent sensor signal origins. \textbf{Left}: original multi-echo point sets. $[\textcolor{red}{\nth{1}}, \textcolor{mygreen}{\nth{2}}, \textcolor{blue}{\nth{3}}]$ echo points. \textbf{Right}: After reassignment. [\textcolor{red}{penetrable}, \textcolor{blue}{impenetrable}] points.} 
\label{fig:reassign}
\vspace{-0.2cm}
\end{figure}

As a result, we present a grouping method to reassign multi-echo points into two new sets. As shown in Fig~\ref{fig:reassign}, the farthest point to the sensor of each `echo group' is assigned to one set and the rest points are assigned to the other set. We call the two new sets `penetrable' and `impenetrable' point set, in the sense that if an echo is not the furthest point in an `echo group', it means the laser can `penetrate' this object and be reflected by objects further away. If a point is assigned to the penetrable set, it is likely that the point is on the contour of an object (partial signal keeps propagating forward), or reflected by a semi-transparent surface (partial signal travels through the surface). It is clear that contour information helps better locate the object, and semi-transparent information encodes the existence of certain part of an object such as the window of a car, both are useful for object box refinement. 

\vspace{1.5mm}
\noindent \textbf{Multi-Echo Aggregation.} 
Given two new sets of points, we then aim to learn a regional RoI feature for accurate bounding box refinement. Since points in two sets encode different object information, we learn two separate features for each set with MLP followed by a point-wise pooling \cite{qsm17pointnet}. Then we form a joint feature of the RoI region by aggregating two feature vectors together. We explore multiple aggregation schemes and choose concatenation as the final method. The refinement network finally adopts a 2-layer MLP which diverges into 2 branches to perform confidence estimation and proposal regression. 

\subsection{Loss Function}
Our proposed method is trained with a multi-task loss, including proposal generation loss $L_{\text{pg}}$ and bounding box refinement loss $L_{\text{refine}}$:
\begin{align}
    L_{\text{overall}} = L_{\text{pg}} + L_{\text{refine}}.
\end{align}

Following \cite{swl19pointrcnn, ysl19std}, the proposal generation loss $L_{\text{pg}}$ consists of a point cloud binary segmentation loss and the proposal regression loss:
\begin{align}
    \mathcal{L}_{\mathrm{pg}} &= \mathcal{L}_{\mathrm{reg}} + \mathcal{L}_{\mathrm{focal}},
\end{align}
where $\mathcal{L}_{\mathrm{focal}}$ is the focal loss used to learn point cloud foreground segmentation as described in \cite{lin2017focal, swl19pointrcnn}. Using a bottom-up bin-based proposal generation module, the proposal regression loss $\mathcal{L}_{\mathrm{reg}}$ is composed of a bin classification loss $L_{\text{bin}}$ and a size residual loss $L_{\text{res}}$. Given the proposal parameter $(x,y,z,h,w,l,\theta)$ where $(x,y,z)$ is the object center, $(h,w,l)$ is the object size and $\theta$ is the orientation, the loss terms are formulated as:
\begin{align}
\begin{aligned}
    \mathcal{L}_{\text {bin}}^{p} &= \sum_{u \in\{x, z, \theta\}}(\mathcal{L}_{\text {cls}}(\widehat{\operatorname{bin}_{u}^{p}}, \operatorname{bin}_{u}^{p})+\mathcal{L}_{\text {L1}}(\widehat{\operatorname {res}_{u}^{p}}, \operatorname{res}_{u}^{p})), \\
    \mathcal{L}_{\text {res}}^{p} &= \sum_{u \in\{y, h, w, l\}}\mathcal{L}_{\text {L1}}(\widehat{\operatorname {res}_{v}^{p}}, \operatorname{res}_{v}^{p}), \\
    \mathcal{L}_{\mathrm{reg}} &= \frac{1}{N_{\mathrm{pos}}} \sum_{p \in \mathrm{pos}}\left(\mathcal{L}_{\mathrm{bin}}^{p}+\mathcal{L}_{\mathrm{res}}^{p}\right),
\end{aligned}
\end{align}
where $\mathcal{L}_{\text {cls}}$ is the classification cross-entropy loss, and $\mathcal{L}_{\text {L1}}$ is the smooth L1 regression loss. $\widehat{\text{bin}^{p}}$ and $\widehat{\operatorname{res}^{p}}$ are predicted bin selection and parameter residual for point $p$, while $\operatorname{bin}^{p}$ and $\operatorname{res}^{p}$ are the ground truth ones. $N_{\text{pos}}$ is the number of total foreground points, so the proposal regression loss is the average sum of bin loss and residual loss for all foreground points. Also, the bounding box refinement loss $L_{\text{refine}}$ is composed of a classification loss for confidence estimation and a regression loss for similar to the previous stage. 
\begin{align}
\begin{aligned}
    \mathcal{L}_{\mathrm{refine}} =& \frac{1}{N_{\mathrm{a}}} \sum_{i}^{N_{\mathrm{a}}}\mathcal{L}_{\mathrm{cls}}(\text{score}_i, \text{label}_i) \\
    &+ \frac{1}{N_{\mathrm{p}}} \sum_{i}^{N_{\mathrm{p}}}(\tilde{\mathcal{L}}_{\text {bin}}^{i} + \tilde{\mathcal{L}}_{\text {res}}^{i}),
\end{aligned}
\end{align}
where $N_{\mathrm{a}}$ is the number of anchor boxes, and $N_{\mathrm{p}}$ is the number of positive proposals for regression, $\mathrm{score}_i$ and $\mathrm{label}_i$ are predicted and ground truth confidence label. $\tilde{\mathcal{L}}_{\text {bin}}^{i}$ and $\tilde{\mathcal{L}}_{\text {res}}^{i}$ are bin and residual loss similar to previous stage, except both the predicted and ground truth bounding box parameters are transformed into the canonical coordinate.

\begin{table*}[!th]
\centering
\caption{\small \textbf{Performance comparison of 3D object detection with state-of-the-art methods on the real-world dataset}. The evaluation metric is Average Precision (AP) with different IoU thresholds.}
\vspace{0.05cm}
\resizebox{\textwidth}{!}{
\begin{tabu}{@{}l|l|ccc|ccc|ccc|ccc@{}}
\toprule
\multicolumn{2}{c|}{\multirow{2}{*} {\text { Method }}} & \multicolumn{3}{c|} {Car - $\text{IoU}=0.7$} & \multicolumn{3}{c|} {Car - $\text{IoU}=0.5$} & \multicolumn{3}{c|} {Person - $\text{IoU}=0.5$} & \multicolumn{3}{c} {Person - $\text{IoU}=0.25$} \\
\multicolumn{2}{c|}{} & Easy & Moderate & Hard & Easy & Moderate & Hard & Easy & Moderate & Hard & Easy & Moderate & Hard \\
\midrule
\multirow{2}{*} {\text { SECOND \cite{yml18second}}} & \nth{1} echo & 67.9 & 37.1 & 27.3 & 79.9 & 57.0 & 56.2 & 42.1 & 25.2 & 19.8 & 54.8 & 35.5 & 23.9\\
& full echo & 75.0 & 42.9 & 30.6 & 86.8 & 65.5 & 65.1 & 47.4 & 29.9 & 20.3 & 58.5 & 39.2 & 25.6\\ 
\midrule
\multirow{2}{*} {\text { PointRCNN \cite{swl19pointrcnn}}} & \nth{1} echo & 66.9 & 37.8 & 28.1 & 80.2 & 57.7 & 52.1 & 45.2 & 28.9 & 20.0 & 57.0 & 38.7 & 25.2 \\ 
& full echo   & 73.6 & 41.9 & 28.9 & 85.0 & 65.6 & 62.7 & 51.6 & 31.4 & 20.2 & 61.2 & 40.7 & 25.9\\ 
\midrule
\multirow{2}{*} {\text { {3DSSD} \cite{ysl203dssd}}} & \nth{1} echo & {64.1} & 36.7 & 27.0 & 77.4 & 55.9 & 52.4 & 45.7 & 28.8 & 20.2 & 57.5 & 38.5 & 23.8\\
& full echo & 72.4 & 40.6 & 28.1 & 83.9 & 65.1 & 63.9 & {51.9} & {30.7} & {19.9} & {60.8} & {40.5} & {26.2}\\ 
\midrule
\multirow{2}{*} {\text { {SASSD} \cite{hzh20structure}}} & \nth{1} echo & 68.8 & 37.9 & 28.6 & 81.2 & 58.3 & 56.8 & 42.5 & 26.7 & 18.2 & 54.7 & 36.6 & 23.2\\
& full echo & 76.1 & 43.2 & 29.8 & 87.2 & 66.0 & 63.7 & {46.9} & {28.5} & {19.4} & {58.1} & {38.6} & {23.7}\\ 
\midrule
\multirow{2}{*} {\text { {PV-RCNN} \cite{sgj20pvrcnn}}} & \nth{1} echo & 69.1 & 38.3 & 28.4 & 81.7 & 59.1 & 57.4 & 44.9 & 28.2 & 19.7 & 56.3 & 38.2 & 24.9\\
& full echo & 76.9 & 44.1 & \textbf{31.2} & 88.1 & 67.2 & 65.3 & 52.7 & 30.9 & 20.8 & 62.0 & 41.2 & 26.2\\ 
\midrule
\multicolumn{2}{c|} {\textbf{MSLiD}} & \textbf{79.5} & \textbf{45.3} & {30.7} & \textbf{89.7} & \textbf{68.1} & \textbf{65.3} & \textbf{57.5} & \textbf{34.2} & \textbf{21.5} & \textbf{66.5} & \textbf{43.2} & \textbf{27.7}\\ 
\multicolumn{2}{c|} {\textit{Improvement}} & +\textit{2.6} & +\textit{1.2} & -\textit{0.5} & +\textit{1.6} & +\textit{0.9} & +\textit{0.0} & +\textit{4.8} & +\textit{2.8} & +\textit{0.7} & +\textit{4.5} & +\textit{2.0} & +\textit{1.5}\\
\bottomrule
\end{tabu}} 
\label{tab:results}
\vspace{-2mm}
\end{table*}

\section{Experiments}

In this section we introduce the dataset (Sec~\ref{sec:dataset}) we used to train and test our approach. We also introduce the implementation details including network architecture and training parameters (Sec~\ref{sec:implementation_details}) used in our experiment. Then we compare our results with other state-of-the-art 3D detection methods on two datasets (Sec~\ref{sec:results}, Sec~\ref{sec:results_sim}) and conduct extensive ablation studies (Sec~\ref{sec:ablation}) to investigate each component of our approach and validate our design choices. 

\subsection{Dataset} \label{sec:dataset}
Since there is no publicly available Multi-signal LiDAR benchmark dataset with ambient illumination, multi-echo point cloud and \intensity measurements for 3D object detection evaluation, we collect two new datasets with the multi-signal measurements to evaluate our method.

\vspace{2mm}
\noindent \textbf{Real Dataset} is collected by a roof-mounted prototype LiDAR on top of a vehicle driving around a North America city. The LiDAR provides three-echo point cloud along with ambient and \intensity value for each point. Each echo group $EG=[\left(p_1, i_1\right),\left(p_2, i_2\right), \left(p_3, i_3\right), a]$, where $p=(x,y,z)$ is a 3D point coordinate, $i$ is \intensity value for each echo point and $a$ is the ambient value for the echo group. The echoes are ordered by signal strength, so the first echo has the highest \intensity value within the echo group. None detected echoes are marked empty with zero \intensity. The converted `LiDAR image' has resolution $[96,600]$, meaning that there are 96 vertically aligned echo groups each column and 600 horizontally aligned echo groups each row. The dataset consists of 35,850 frames collected from various driving scenes including downtown, highway, suburban areas, etc. We split our dataset into training and testing sets with a 70/30 ratio, where the training set consists of 25,002 frames and validation set has 10,848 frames, both sampled from different video clips with high diversity. Ground truth labels of the dataset are 3D oriented bounding boxes of `Person' and `Car' classes in the 3D space, and the 2D bounding boxes of the same classes in the `LiDAR image' space. 

\vspace{2mm}
\noindent \textbf{Synthetic Dataset} is collected using the CARLA \cite{drc17carla} simulator. It is a large-scale multi-sensor multi-task dataset. Similar to the real dataset, we also collect three-echo point cloud with \intensity value. We approximate ambient value with r-channel of RGB image because they both capture ambient sunlight signal of close wavelength. We use 26,043 frames for training and 8,682 frames for testing. The format of the synthetic dataset is the same as the real dataset. Moreover, the synthetic dataset provides a wide range of ground truth annotation including 2D and 3D bounding boxes and segmentation. Among various classes, we focus on `Car', `Person' and `Cyclist.' We will release our synthetic dataset to public for reproduction and competition. More details of our synthetic dataset are provided in the supplementary.

\subsection{Implementation Details} \label{sec:implementation_details}
\noindent \textbf{Network Architecture.}\ \ In order to align network inputs, we randomly subsample 16K points from multi-echo point cloud in each scene. Note that multiple point clouds are treated as one whole set to sample from. In the proposal generation stage, we follow the network structure of \cite{qys17pointnet++} with four set abstraction ($[4096,1024,256,64]$ with multi-scale grouping) and four feature propagation layers as our 3D feature extraction backbone. For 2D detector, we use Faster-RCNN \cite{rhg15fasterrcnn} with Feature Pyramid Networks (FPN) \cite{lpg17fpn} module and ResNet-50 \cite{hzr16resnet} as backbone. 

In the bounding box refinement network, we randomly sample 256 points from each reassigned point set as input to MEA module. We follow the network structure of \cite{qys17pointnet++} with three set abstraction layers ([64, 16, 1]) to generate a single feature vector for each of the point sets. Then the two features are concatenated to jointly perform estimation and regression heads.

\vspace{2mm}
\noindent \textbf{Training Parameters.}\ \ The two stages of our methods are trained separately using Adam optimizer \cite{kb14adam}. For real-world dataset, stage-1 is trained for 150 epoch with learning rate 0.002, and stage-2 is trained for 50 epochs with learning rate 0.001. For synthetic data, stage-1 is trained for 100 epoch with learning rate 0.002, and stage-2 is trained for 40 epochs with learning rate 0.002. For bin-based proposal generation and refinement module, we adopt the same bin size, search range and orientation numbers as in \cite{swl19pointrcnn}. During confidence estimation, a `Car' proposal is considered positive if its maximum 3D IoU is above 0.6, and negative if its below 0.45. For `Person', the positive and negative thresholds are set to be 0.5 and 0.4.

Our 2D detector is pre-trained on ImageNet classification and COCO object detection. We then fine-tune it on our datasets using LiDAR image as inputs to detect 2D bounding boxes. Batch size, weight decay and momentum are set to be 8, 1e-4 and 0.9 for both datasets. Learning rate is set to be 0.005 for real-world dataset and 0.01 for synthetic dataset. Horizontal flipping is used for data augmentation. 

\subsection{Results on Real-World Dataset} \label{sec:results}
We compare our model against state-of-the-art point-based and grid-based 3D object detectors under different point sets. For evaluation metric we use average precision (AP) under different IoU thresholds, where for `Car' class we use $\mathrm{IoU}=\{0.5, 0.7\}$, and for `Person' we use $\mathrm{IoU}=\{0.25, 0.5\}$. Difficulty level is chosen based on the depth value, where easy class contains objects within 40$m$, moderate class contains objects between 40-80$m$ and hard class contains objects between 80-200$m$. 

\vspace{2mm}
\noindent \textbf{Comparison with state-of-the-art methods.}\ \ We show the evaluation results and comparison with SOTA methods in Table~\ref{tab:results}. Note that rows corresponding to ``\nth{1} echo'' refers to experiments using the strongest echo, i.e., with the highest intensities -- This mimics the classic single-echo point cloud data where only the strongest point is preserved. Rows corresponding to full echo means ``multi-echo points grouped as one point cloud.'' On both `Car' and `Person' class with different IoU threshold, our method outperforms state-of-the-art methods with remarkable margins. For `Car' class, our method achieves up to 2.9 AP increase and for `Person' class we achieve up to 4.8 improvement. The improvements over `Easy' and `Moderate' classes are both noticeable. Note that we do not get noticeable improvement for hard objects farther than 80$m$, because multiple modalities provide less information faraway objects. First, the number of multi-echo points decreases quickly as distance increases because of the signal attenuation. Also, they are hard to detect on LiDAR image because of the small size. Under `Person' class, our method outperforms previous methods on all three difficulty levels.

Notice that all previous methods using ``full echo'' perform better than using only ``\nth{1} echo'', despite that this is achieved by simply merging points from available echos together as a single point cloud. This means that raw information contained in other echos (e.g., \nth{2}, \nth{3} echos) in addition to the strongest echo (i.e., the \nth{1} echo) can be used to improve performance. Moreover, our method is designed to better extract and aggregate features from multi-signal information, so can achieve even higher performance.

\begin{table}[!t]
\centering
\caption{Effects of different input signals on overall 3D detection performance, where SE stands for single-echo and ME stands for multi-echo.}
\vspace{0.05cm}
\resizebox{0.47\textwidth}{!}{
\begin{tabu}{@{}c|c|c||c@{}}\hline
Point Cloud & {Ambient Signal} & {ME Reflectance Signal} & Overall AP \\\hline\hline
SE &  &  & 50.5 \\
ME &  &  & 53.6 \\
ME & \checkmark &  & 53.7 \\
ME &  & \checkmark & 55.1 \\
ME & \checkmark & \checkmark & \textbf{55.5} \\\hline
\end{tabu}
} 
\label{tab:ablation_input}
\end{table}

\begin{table}[!t]
\centering
\caption{Effects of our proposed components on overall detection, including the MSF module and MEA module.}
\vspace{0.05cm}
\resizebox{0.47\textwidth}{!}{
\begin{tabular}{@{}c|c|c|c||c@{}}\hline
\multirow{2}{*}{Method} & \multicolumn{2}{c|}{MSF Module} & \multirow{2}{*}{MEA module} & \multirow{2}{*}{Overall AP} \\\cline{2-3}
& {class feature} & {pixel feature} &  & \\\hline\hline
baseline  &  &  &   & 50.5  \\
MSF only & \checkmark & \checkmark &  & 52.1 \\
MEA only &  & & \checkmark & 53.6 \\
w/o class feature &  & \checkmark & \checkmark & 54.0 \\
w/o pixel feature & \checkmark &  & \checkmark & 55.3 \\
\textbf{MSLiD} & \checkmark & \checkmark & \checkmark & \textbf{55.5} \\\hline
\end{tabular}
} 
\label{tab:ablation_proposed}
\end{table}

\begin{table*}[!t]
\centering
\caption{\small \textbf{Performance on synthetic dataset}. Previous SOTA methods are trained with all multi-echo points grouped as one (Full echo training). The evaluation metric is Average Precision (AP).}
\vspace{0.05cm}
\resizebox{0.85\textwidth}{!}{
\begin{tabular}{@{}c|ccc|ccc|ccc@{}} 
\toprule
\multirow{2}{*} {\text { Method }} & \multicolumn{3}{c|} {Car - $\text{IoU}=0.7$} & \multicolumn{3}{c|} {Pedestrian - $\text{IoU}=0.5$} & \multicolumn{3}{c} {Cyclist - $\text{IoU}=0.5$} \\
& Easy & Moderate & Hard & Easy & Moderate & Hard & Easy & Moderate & Hard \\
\midrule
\text { SECOND \cite{yml18second}} & 76.9 & 70.8 & 63.2 & 56.2 & 51.8 & 43.1 & 60.7 & 55.5 & 48.7 \\
\text { PointRCNN \cite{swl19pointrcnn}} & 77.8 & 71.4 & 63.3 & 59.9 & 53.4 & 44.8 & 61.1 & 55.7 & 48.8 \\
\text { PV-RCNN \cite{sgj20pvrcnn}} & 80.4 & 72.8 & 64.5 & 61.2 & 54.7 & 45.5 & 63.5 & 57.0 & 49.3 \\
\midrule
{\textbf{MSLiD}} & \textbf{81.6} & \textbf{73.9} & \textbf{66.9} & \textbf{66.2} & \textbf{59.1} & \textbf{49.3} & \textbf{64.6} & \textbf{58.2} & \textbf{51.2} \\
{\textit{Improvement}} & +\textit{1.4} & +\textit{1.1} & +\textit{2.4} & +\textit{5.0} & +\textit{4.4} & +\textit{3.8} & +\textit{1.1} & +\textit{1.2} & +\textit{1.9} \\
\bottomrule
\end{tabular}
} 
\vspace{0.1mm}
\label{tab:results_sim}
\end{table*}


\subsection{Ablation Study} \label{sec:ablation}
We conduct extensive ablation experiments to analyze the effectiveness of different proposed components of our model and other design choices. Following \cite{zto18voxelnet, ysl19std}, all ablation studies are conducted on the `Car' class.

\vspace{2mm}
\noindent \textbf{Effect of different input signal.} In Table~\ref{tab:ablation_input}, we first test the point cloud sets where we take only single-echo (SE) or multi-echo (ME) point cloud as input. For ME we use three echo groups of points and for SE, we merge three groups together as one single group. The first two rows shows that our method which leverages multi-echo point features through the MEA module is better than simply merging all echo groups, achieving a 3-point improvement in AP. We then ablate the ambient signal as well as the ME \intensity signal to show their effect. As the table indicates, MSLiD also makes improvement on detection results by properly leveraging these two types of LiDAR signals -- Adding ambient and ME \intensity signal improves the overall AP by 1.9\% from the point-cloud-only baseline.

\vspace{2mm}
\noindent \textbf{Effect of MSF and MEA modules.} In Table~\ref{tab:ablation_proposed} the \nth{1} row, we remove both the MSF module (use 3D GNN feature only) and the MEA module (treats all points identically). This results in a baseline working with only single-echo point cloud. In the following two rows, we disable one of the two modules to validate how it contributes to the our method. Then, we look inside MSF module and ablate each of the two 2D branch feature vectors. As table shows, the two proposed modules together improve absolute AP by 5.0 from the baseline. For the `Car' class, multi-echo aggregation tends to contribute more between the two modules. This improvement comes from that multi-echo points encodes contour and surface reflectivity information useful to estimate location, size and orientation of the object. 


\begin{table}[!t]
\centering
\caption{Performance for difference feature aggregation schemes (column 1-3) and point cloud set definition (column 4-5) of bounding box refinement stage. $pc^{\mathrm{echo}}$ represents the raw multi-echo point cloud, and $pc^{\mathrm{reassign}}$ represents point sets after using reassignment strategy described in Sec~\ref{sec:reassign}.}
\vspace{0.05cm}
\resizebox{0.485\textwidth}{!}{
\begin{tabular}{@{}ccc|cc|c@{}}
\toprule
\multicolumn{3}{c|} {Aggregation Scheme} & \multicolumn{2}{c|} {Point Cloud Sets} & \multirow{2}{*} {Overall AP} \\\cmidrule{1-5}
{Max-P}  & {Mean-P}  & {Concat.} & $\text{pc}^{\mathrm{echo}}$ & $\text{pc}^{\mathrm{reassign}}$ &  \\
\midrule
\checkmark &  &  &  \checkmark &  & 51.1\\
& \checkmark & &  \checkmark & & 53.9\\
& & \checkmark & \checkmark & & 54.5\\
& & \checkmark & & \checkmark & \textbf{55.5}\\
\bottomrule
\end{tabular}
} 
\label{tab:ablation_mea}
\end{table}

\vspace{1mm}
\noindent \textbf{Effect of point cloud reassignment and aggregation.} 
We show the results in Table~\ref{tab:ablation_mea}, where Max-P, Mean-P and Concat represent different methods to aggregate features learned from different point sets, $pc^{\mathrm{echo}}$ and $pc^{\mathrm{reassign}}$ represent different point sets definition. The overall AP is calculated over all ground truth proposals without considering difficulty levels. From the \nth{1}-\nth{3} row we can see that using concatenation to aggregate features of different sets results in the best performance, while max-pooling tends to get low performance. This performance gap is because features learned from multiple point sets encode complementary information, which can be better encoded by concatenation than pooling. From the last two rows, we can see that our point set reassignment strategy helps better learn the RoI feature and further improve the absolute AP of 1.

\subsection{Results on Synthetic Dataset} \label{sec:results_sim}
We also validate our method on the synthetic dataset. For the synthetic dataset, the difficulty level follows the same definition as in the KITTI benchmark \cite{glu12kitti}, where easy, moderate and hard are arranged by 2D bounding box size and occlusion/truncation level. For `Car', we use $\mathrm{IoU}=0.7$, and for `Pedestrian' and `Cyclist' we use $\mathrm{IoU}=0.5$. As shown in Table~\ref{tab:results_sim}, our proposed method outperforms 
state-of-the-art methods by large margins on all three classes. For `Car' and `Cyclist' classes, we get even higher improvements on the hard class, because the hard class is not defined merely by distance, so multi-signal information still helps better detect truncated/occluded objects. The experiment further proves that our proposed method has a good generalization property and works better for different datasets. 

\subsection{Discussion on Cost-Benefit Trade-off}
The size of MSLiD is 10G with a batch size of 4. The inference time is 110ms on a RTX 2080Ti GPU, with PointNet++\cite{qys17pointnet++} backbone taking bulk of the time. We also show that leveraging multi-echo points for performance improvements does not significantly increase the running time. In our experiments, we always sample a total of 16K points in different settings (full echo or only the 1st echo). Thus, adding the MSF module or not requires the same time to process the full echo or only the 1st echo. The only source of additional time cost when using multiple echoes in our method comes from the MEA module, where a 3-layer MLP is used to learn separate RoI features for two point sets. It increases AP by 3.4 according to Table~\ref{tab:ablation_proposed} with marginal increase of inference time. Also, there is no additional space needed for multi-echo fusion as we sample the same number of points in different settings.

\section{Conclusion}
We proposed the first method exploring to fuse a wide range of Multi-signal LiDAR information for 3D object detection. Our method takes advantage of multi-echo point clouds and ambient/\intensity signals to learn discriminative point features and proposal regional RoI features. In the proposal generation stage, a multi-signal fusion (MSF) module was proposed to fuse 2D CNN features learned from `LiDAR image' with 3D GNN features learned from point cloud. In the refinement stage, a multi-echo aggregation (MEA) module was proposed to learn a better object context RoI feature from multi-echo point clouds. The better RoI feature leads to accurate bounding box refinement.  
Our proposed method achieved state-of-the-art performance in two datasets with multiple LiDAR measurements.

{\small
\bibliographystyle{ieee_fullname}
\bibliography{main}

\begin{thebibliography}{10}\itemsep=-1pt

\bibitem{Caccia2019}
Lucas Caccia, {Herke Van Hoof}, Aaron Courville, and Joelle Pineau.
\newblock {Deep Generative Modeling of LiDAR Data}.
\newblock {\em IROS}, 2019.

\bibitem{cmw17mv3d}
Xiaozhi Chen, Huimin Ma, Ji Wan, Bo Li, and Tian Xia.
\newblock Multi-view 3d object detection network for autonomous driving.
\newblock In {\em Proceedings of the IEEE Conference on Computer Vision and
  Pattern Recognition}, pages 1907--1915, 2017.

\bibitem{drc17carla}
Alexey Dosovitskiy, German Ros, Felipe Codevilla, Antonio Lopez, and Vladlen
  Koltun.
\newblock Carla: An open urban driving simulator.
\newblock {\em arXiv preprint arXiv:1711.03938}, 2017.

\bibitem{glu12kitti}
Andreas Geiger, Philip Lenz, and Raquel Urtasun.
\newblock Are we ready for autonomous driving? the kitti vision benchmark
  suite.
\newblock In {\em 2012 IEEE Conference on Computer Vision and Pattern
  Recognition}, pages 3354--3361. IEEE, 2012.

\bibitem{gev18sparse}
Benjamin Graham, Martin Engelcke, and Laurens Van Der~Maaten.
\newblock 3d semantic segmentation with submanifold sparse convolutional
  networks.
\newblock In {\em Proceedings of the IEEE conference on computer vision and
  pattern recognition}, pages 9224--9232, 2018.

\bibitem{hzh20structure}
Chenhang He, Hui Zeng, Jianqiang Huang, Xian-Sheng Hua, and Lei Zhang.
\newblock Structure aware single-stage 3d object detection from point cloud.
\newblock In {\em Proceedings of the IEEE/CVF Conference on Computer Vision and
  Pattern Recognition}, pages 11873--11882, 2020.

\bibitem{hzr16resnet}
Kaiming He, Xiangyu Zhang, Shaoqing Ren, and Jian Sun.
\newblock Deep residual learning for image recognition.
\newblock In {\em Proceedings of the IEEE conference on computer vision and
  pattern recognition}, pages 770--778, 2016.

\bibitem{hlc20epnet}
Tengteng Huang, Zhe Liu, Xiwu Chen, and Xiang Bai.
\newblock Epnet: Enhancing point features with image semantics for 3d object
  detection.
\newblock In {\em European Conference on Computer Vision}, pages 35--52.
  Springer, 2020.

\bibitem{kb14adam}
Diederik~P Kingma and Jimmy Ba.
\newblock Adam: A method for stochastic optimization.
\newblock {\em arXiv preprint arXiv:1412.6980}, 2014.

\bibitem{kml18avod}
Jason Ku, Melissa Mozifian, Jungwook Lee, Ali Harakeh, and Steven~L Waslander.
\newblock Joint 3d proposal generation and object detection from view
  aggregation.
\newblock In {\em 2018 IEEE/RSJ International Conference on Intelligent Robots
  and Systems (IROS)}, pages 1--8. IEEE, 2018.

\bibitem{lvc19pointpillars}
Alex~H Lang, Sourabh Vora, Holger Caesar, Lubing Zhou, Jiong Yang, and Oscar
  Beijbom.
\newblock Pointpillars: Fast encoders for object detection from point clouds.
\newblock In {\em Proceedings of the IEEE Conference on Computer Vision and
  Pattern Recognition}, pages 12697--12705, 2019.

\bibitem{lyc19mtms}
Ming Liang, Bin Yang, Yun Chen, Rui Hu, and Raquel Urtasun.
\newblock Multi-task multi-sensor fusion for 3d object detection.
\newblock In {\em Proceedings of the IEEE Conference on Computer Vision and
  Pattern Recognition}, pages 7345--7353, 2019.

\bibitem{lyw18continuous}
Ming Liang, Bin Yang, Shenlong Wang, and Raquel Urtasun.
\newblock Deep continuous fusion for multi-sensor 3d object detection.
\newblock In {\em Proceedings of the European Conference on Computer Vision
  (ECCV)}, pages 641--656, 2018.

\bibitem{lpg17fpn}
Tsung-Yi Lin, Piotr Doll{\'a}r, Ross Girshick, Kaiming He, Bharath Hariharan,
  and Serge Belongie.
\newblock Feature pyramid networks for object detection.
\newblock In {\em Proceedings of the IEEE conference on computer vision and
  pattern recognition}, pages 2117--2125, 2017.

\bibitem{lin2017focal}
Tsung-Yi Lin, Priya Goyal, Ross Girshick, Kaiming He, and Piotr Doll{\'a}r.
\newblock Focal loss for dense object detection.
\newblock In {\em Proceedings of the IEEE international conference on computer
  vision}, pages 2980--2988, 2017.

\bibitem{sdh20clocs}
S. {Pang}, D. {Morris}, and H. {Radha}.
\newblock Clocs: Camera-lidar object candidates fusion for 3d object detection.
\newblock In {\em 2020 IEEE/RSJ International Conference on Intelligent Robots
  and Systems (IROS)}, pages 10386--10393, 2020.

\bibitem{qcl20imvotenet}
Charles~R Qi, Xinlei Chen, Or Litany, and Leonidas~J Guibas.
\newblock Imvotenet: Boosting 3d object detection in point clouds with image
  votes.
\newblock In {\em Proceedings of the IEEE/CVF Conference on Computer Vision and
  Pattern Recognition}, pages 4404--4413, 2020.

\bibitem{qlw18fpointnet}
Charles~R Qi, Wei Liu, Chenxia Wu, Hao Su, and Leonidas~J Guibas.
\newblock Frustum pointnets for 3d object detection from rgb-d data.
\newblock In {\em Proceedings of the IEEE conference on computer vision and
  pattern recognition}, pages 918--927, 2018.

\bibitem{qsm17pointnet}
Charles~R Qi, Hao Su, Kaichun Mo, and Leonidas~J Guibas.
\newblock {PointNet: Deep Learning on Point Sets for 3D Classification and
  Segmentation}.
\newblock {\em CVPR}, 2017.

\bibitem{qys17pointnet++}
Charles~R Qi, Li Yi, Hao Su, and Leonidas~J Guibas.
\newblock {PointNet++: Deep Hierarchical Feature Learning on Point Sets in a
  Metric Space}.
\newblock {\em NeurIPS}, 2017.

\bibitem{rhg15fasterrcnn}
Shaoqing Ren, Kaiming He, Ross Girshick, and Jian Sun.
\newblock Faster r-cnn: Towards real-time object detection with region proposal
  networks.
\newblock In {\em Advances in neural information processing systems}, pages
  91--99, 2015.

\bibitem{sgj20pvrcnn}
Shaoshuai Shi, Chaoxu Guo, Li Jiang, Zhe Wang, Jianping Shi, Xiaogang Wang, and
  Hongsheng Li.
\newblock Pv-rcnn: Point-voxel feature set abstraction for 3d object detection.
\newblock In {\em Proceedings of the IEEE/CVF Conference on Computer Vision and
  Pattern Recognition (CVPR)}, June 2020.

\bibitem{swl19pointrcnn}
Shaoshuai Shi, Xiaogang Wang, and Hongsheng Li.
\newblock Pointrcnn: 3d object proposal generation and detection from point
  cloud.
\newblock In {\em Proceedings of the IEEE Conference on Computer Vision and
  Pattern Recognition}, pages 770--779, 2019.

\bibitem{sws20parta2}
Shaoshuai Shi, Zhe Wang, Jianping Shi, Xiaogang Wang, and Hongsheng Li.
\newblock From points to parts: 3d object detection from point cloud with
  part-aware and part-aggregation network.
\newblock {\em IEEE Transactions on Pattern Analysis and Machine Intelligence},
  2020.

\bibitem{vlh20pointpainting}
Sourabh Vora, Alex~H Lang, Bassam Helou, and Oscar Beijbom.
\newblock Pointpainting: Sequential fusion for 3d object detection.
\newblock In {\em Proceedings of the IEEE/CVF Conference on Computer Vision and
  Pattern Recognition}, pages 4604--4612, 2020.

\bibitem{wj19frustum}
Zhixin Wang and Kui Jia.
\newblock Frustum convnet: Sliding frustums to aggregate local point-wise
  features for amodal.
\newblock In {\em 2019 IEEE/RSJ International Conference on Intelligent Robots
  and Systems (IROS)}, pages 1742--1749. IEEE, 2019.

\bibitem{aiodrive}
Xinshuo Weng, Yunze Man, Dazhi Cheng, Jinhyung Park, Matthew O'Toole, and Kris
  Kitani.
\newblock {All-In-One Drive: A Large-Scale Comprehensive Perception Dataset
  with High-Density Long-Range Point Clouds}.
\newblock {\em arXiv}, 2020.

\bibitem{yml18second}
Yan Yan, Yuxing Mao, and Bo Li.
\newblock Second: Sparsely embedded convolutional detection.
\newblock {\em Sensors}, 18(10):3337, 2018.

\bibitem{ylu18hdnet}
Bin Yang, Ming Liang, and Raquel Urtasun.
\newblock Hdnet: Exploiting hd maps for 3d object detection.
\newblock In {\em Conference on Robot Learning}, pages 146--155, 2018.

\bibitem{ylu18pixor}
Bin Yang, Wenjie Luo, and Raquel Urtasun.
\newblock Pixor: Real-time 3d object detection from point clouds.
\newblock In {\em Proceedings of the IEEE conference on Computer Vision and
  Pattern Recognition}, pages 7652--7660, 2018.

\bibitem{ysl203dssd}
Zetong Yang, Yanan Sun, Shu Liu, and Jiaya Jia.
\newblock 3dssd: Point-based 3d single stage object detector.
\newblock In {\em Proceedings of the IEEE/CVF Conference on Computer Vision and
  Pattern Recognition}, pages 11040--11048, 2020.

\bibitem{ysl19std}
Zetong Yang, Yanan Sun, Shu Liu, Xiaoyong Shen, and Jiaya Jia.
\newblock Std: Sparse-to-dense 3d object detector for point cloud.
\newblock In {\em Proceedings of the IEEE International Conference on Computer
  Vision}, pages 1951--1960, 2019.

\bibitem{ykk203dcvf}
Jin~Hyeok Yoo, Yeocheol Kim, Ji~Song Kim, and Jun~Won Choi.
\newblock 3d-cvf: Generating joint camera and lidar features using cross-view
  spatial feature fusion for 3d object detection.
\newblock {\em arXiv preprint arXiv:2004.12636}, 2020.

\bibitem{zto18voxelnet}
Yin Zhou and Oncel Tuzel.
\newblock Voxelnet: End-to-end learning for point cloud based 3d object
  detection.
\newblock In {\em Proceedings of the IEEE Conference on Computer Vision and
  Pattern Recognition}, pages 4490--4499, 2018.

\bibitem{zmp20cross}
Ming Zhu, Chao Ma, Pan Ji, and Xiaokang Yang.
\newblock Cross-modality 3d object detection.
\newblock {\em arXiv preprint arXiv:2008.10436}, 2020.

\end{thebibliography}
}

\newpage

\appendix

\begin{algorithm*}[t]\caption{Multi-echo LiDAR Simulation}
\algrenewcommand\algorithmicrequire{\textbf{Inputs:}}
\algrenewcommand\algorithmicensure{\textbf{Outputs:}}
\algnewcommand\algorithmicparameters{\textbf{Parameters:}}
\algnewcommand\algparameters{\item[\algorithmicparameters]}
\begin{algorithmic}[1]
\Require \begin{varwidth}[t]{\linewidth}
                  $\mathbf{I_{rgb}}$: RGB image \par
                  $\mathbf{I_{d}}$: Depth image \par
                  $\mathbf{I_{n}}$: Surface normal image \strut
               \end{varwidth}
\vspace{2mm}
\algparameters \begin{varwidth}[t]{\linewidth}
                  $\mathbf{N}$: Number of time bins \par
                  $\mathbf{SBR}$: Signal-Background-Ratio \par
                  $\mathbf{R_H}, \mathbf{F_H}$: Horizontal Resolution and FoV \par
                  $\mathbf{R_V}, \mathbf{F_V}$: Vertical Resolution and FoV \par 
                  $\mathbf{S}$: Neighborhood aggregation kernel size \par 
                  $\mathbf{K}$: Number of echo groups
               \end{varwidth}
\vspace{2mm}
\Ensure \begin{varwidth}[t]{\linewidth}
                  $\mathbf{I_{\ambient}}[R_V, R_H]$: Ambient image \par
                  $\mathbf{I_{\reflectance}}[R_V, R_H, K]$: Reflectance image \par
                  $\mathbf{P_K}$: Multi-echo Point Cloud Returns \strut
               \end{varwidth}

\vspace{4mm}
\State Sample LiDAR sensor array $A[R_H, R_V]$ in the polar coordinate according to FoV $F_H, F_V$\;
\State Project $A[R_H, R_V]$ onto the 2D image space and get the positional map $A_{2D}[R_H, R_V]$
\State Transform the input images $I_{\mathsf{im}}$ into polar coordinate $I_{\mathsf{im}^\prime}$ as Eq.~\ref{eq:interp}\; \Comment{Get ambient image $\mathbf{I_{\ambient}}[R_V, R_H]$}

\vspace{2mm}
\State Calculate signal photons $N_{\mathsf{signal}}$ for each pixel $(h, w)$ from Eq.~\ref{eq:spad_sim1}; 
\State Calculate ambient photons $N_{\mathsf{ambient}}$ for each pixel $(h, w)$ from Eq.~\ref{eq:spad_sim2}; 
\State Calculate ambient photons $N_{\mathsf{p}}[R_V, R_H]$ for each pixel $(h, w)$ from Eq.~\ref{eq:spad_eq}; 

\vspace{2mm}
\State Simulate multi-echo mechanism and generate $\mathbf{N^*_p}[R_H, R_V, N]$ by neighborhood aggregation as in Eq.~\ref{eq:spad_ff}\;
\State Get Top-K returns along the temporal axis with Eq. \ref{eq:spad_top}\;
\State Project the valid bins into 3D space and generate Top-K point cloud returns\;  \Comment{Get multi-echo point cloud $\mathbf{P_K}$}
\State Simulate the reflectance by number of points insides its corresponding bins\;
\State Rearrange reflectance values of points into `LiDAR Image' space \Comment{Get reflectance image $\mathbf{I_{\reflectance}}[R_V, R_H, K]$}

\end{algorithmic}
\label{alg}
\end{algorithm*}

\section*{Supplementary}

\vspace{1mm}
\section{Synthetic Dataset Simulation}
\vspace{-0.1cm}

Though we briefly describe the synthetic multi-signal LiDAR measurement generation process and its data format in the main paper, here we provide details about the entire data simulation process. We will introduce our simulation of ambient signals (Sec~\ref{supp:ambient}) as well as multi-echo point cloud and reflectance signals (Sec~\ref{supp:mepc}).

\subsection{Input Coordinate Transformation}\label{sec:trans}
As shown in Algorithm~\ref{alg}, we take RGB, depth and surface normal images, denotes as $I_{rgb}, I_d, I_n$, as the input of the simulation process, where the RGB and depth signals are directly captured by top-mounted RGBD cameras \cite{aiodrive}, and the surface normal is directly converted from depth images -- The process is accurate because of the perfect synthetic depth values.

In order to mimic the spinning mechanism of LiDAR sensor, we perform a Polar-to-Image coordinate transformation on all input images. Specifically, we approximate the LiDAR sensor as a point in the 3D space and define a LiDAR sensor array $A[R_V, R_H]$ in the polar coordinate (\emph{i.e.} elevation and azimuth). For all $i \in [1, R_V], j \in [1, R_H]$,
\begin{align}
    A(i, j) = (\theta_i, \gamma_j), \ \ \ \ \ \ \theta_i, \gamma_j \in [0, 2\pi)
\end{align}
where $\theta$ and $\gamma$ are vectors representing the LiDAR sensor detector arrangement, and are predefined according to the desired vertical and horizontal FoV and resolution, and $R_H$ denotes the number of vertically aligned LiDAR detectors on the sensor, and $R_V$ denotes the number of horizontally aligned detection a detector performs during one single sweep. Then, we project the polar coordinate map $A[R_V, R_H]$ onto the 2D image space, resulting in non-linearly arranged positional map $A_{2D}[R_V, R_H]$ on the 2D image plane. Then, we generate new images from the input images by sampling and interpolating pixels according to the 2D positional map $A_{2D}[R_V, R_H]$. For any input image $I_{im} \in \{I_{rgb}, I_d, I_n\}$, $\forall i \in [1, R_V], j \in [1, R_H]$,
\begin{align}\label{eq:interp}
    I_{im}^{\prime}(i, j) = \text{interp}(I_{im}(x, y)), \ \ (x, y) =  A_{2D}(i, j)
\end{align}
Then we get new images $I_{rgb}^{\prime}, I_d^{\prime}, I_n^{\prime}$ where pixels are arranged according to the arrangement of LiDAR detector orientations.

\subsection{Ambient Illumination Signals}\label{supp:ambient}
The ambient illumination signals are sunlight reflected by objects. Thus, from the imaging perspective, the information encoded in the ambient signals is very similar to that collected by RGB camera. Collected using CARLA \cite{drc17carla} simulator, the AIODrive dataset \cite{aiodrive} captures RGB images in the scene by deploying RGB cameras on top of the ego-vehicles. Since LiDAR sensor usually captures light with infrared wavelength, we take R-channel from the RGB images and treat it as a simulation of the ambient signal. The ambient signal is represented in the form of images $I_{\ambient}$,
\begin{align*}
    I_{\ambient} \sim I^{\prime}_r
\end{align*}
where $I^{\prime}_r$ denotes the R-channel of image $I^{\prime}$ after coordinate transformation.

\begin{figure*}[t]
    \centering
    \includegraphics[trim=40 218 15 235, clip=true, width=\linewidth]{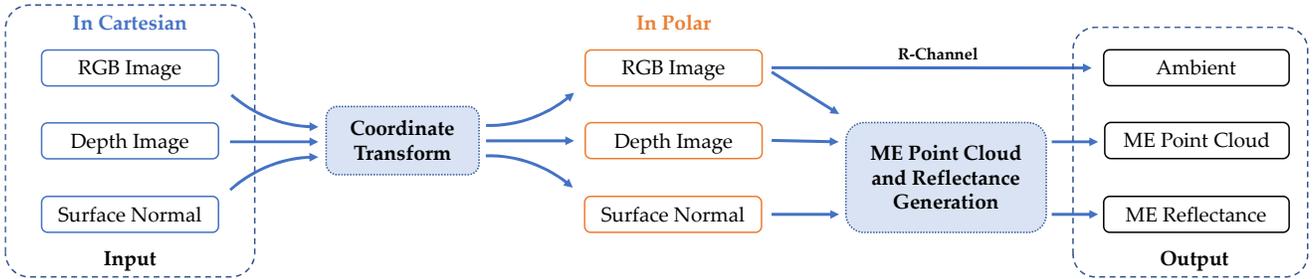}
    \vspace{-0.25cm}
    \caption{\textbf{Multi-signal LiDAR measurements Simulation.} Taking the inputs from AIODrive \cite{aiodrive}, we first transform the images from Cartesian space into polar space to mimic the LiDAR spinning mechanism. Then infrared ambient illumination is approximated by taking R channel of the RGB image. Multi-echo point cloud and reflectance signals are generated from trasformed RGB, Depth and surface normal images.}
    \label{fig:spad_lidar}
    \vspace{-0.3cm}
\end{figure*}

 \begin{figure*}[!t]
    \centering
    \includegraphics[trim=220 340 800 500, clip=true, width=0.327\linewidth]{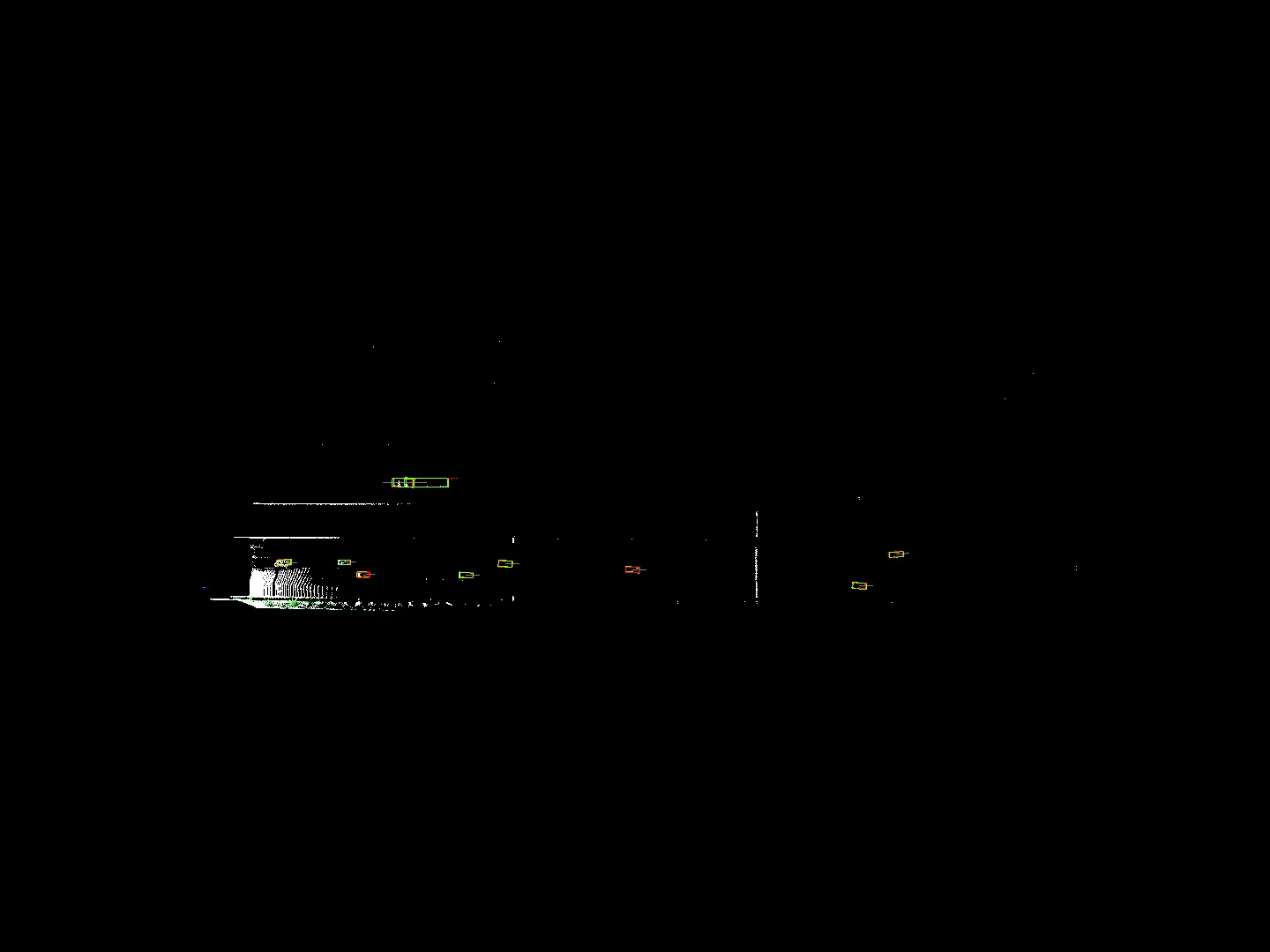}
    \includegraphics[trim=220 340 800 500, clip=true, width=0.327\linewidth]{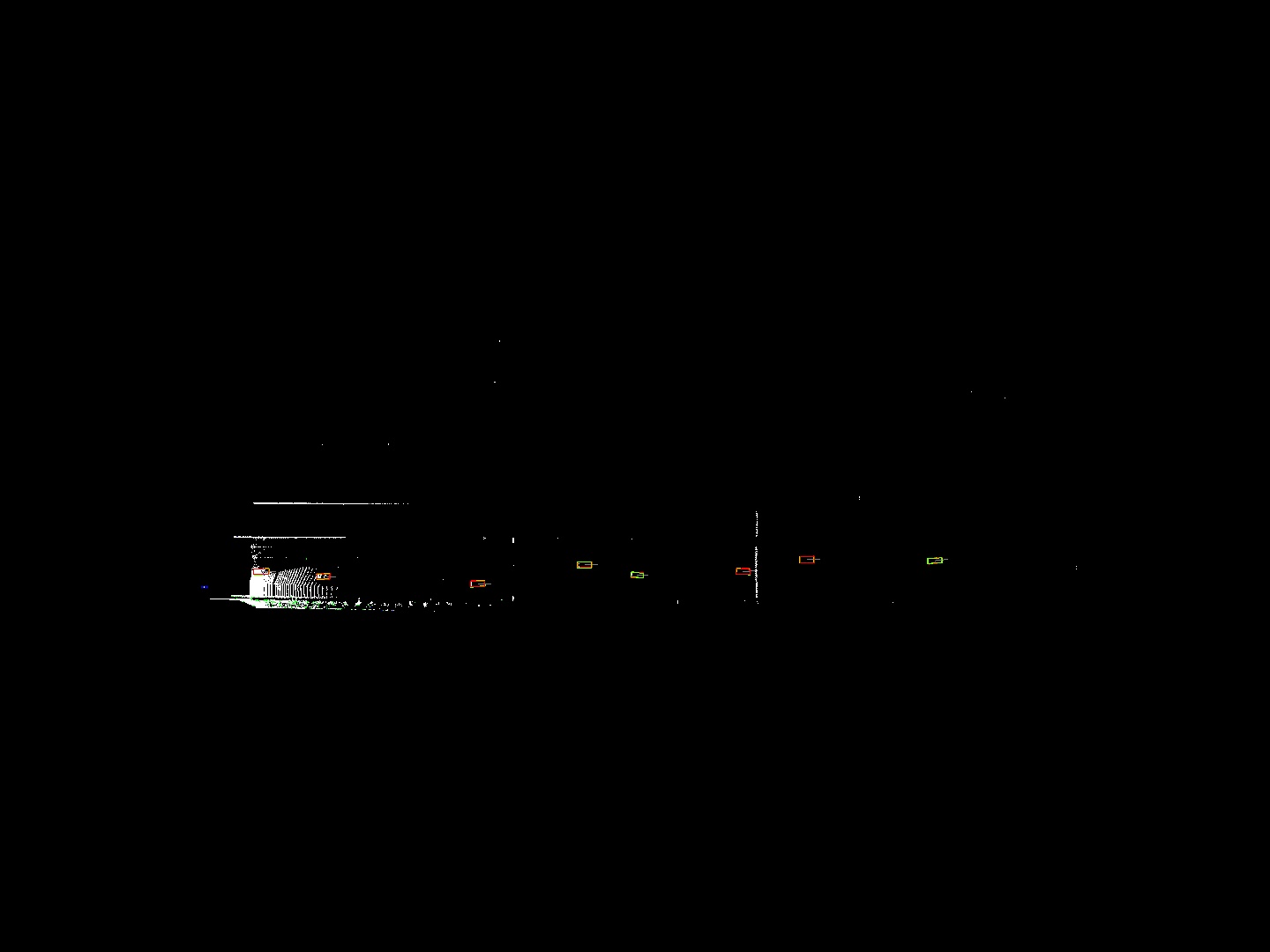}
    \includegraphics[trim=220 340 800 500, clip=true, width=0.327\linewidth]{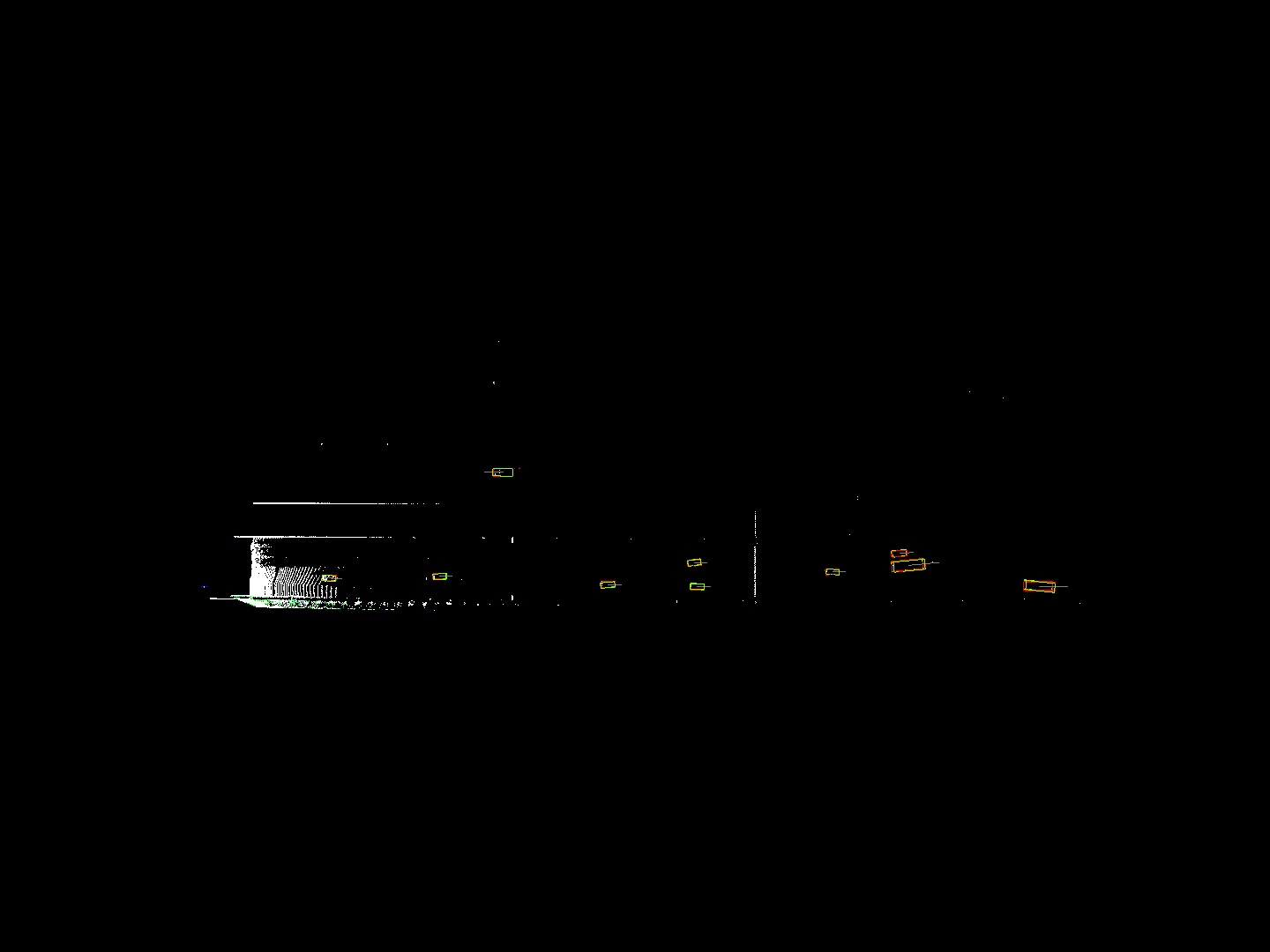} \\
    \includegraphics[trim=220 380 840 350, clip=true, width=0.327\linewidth]{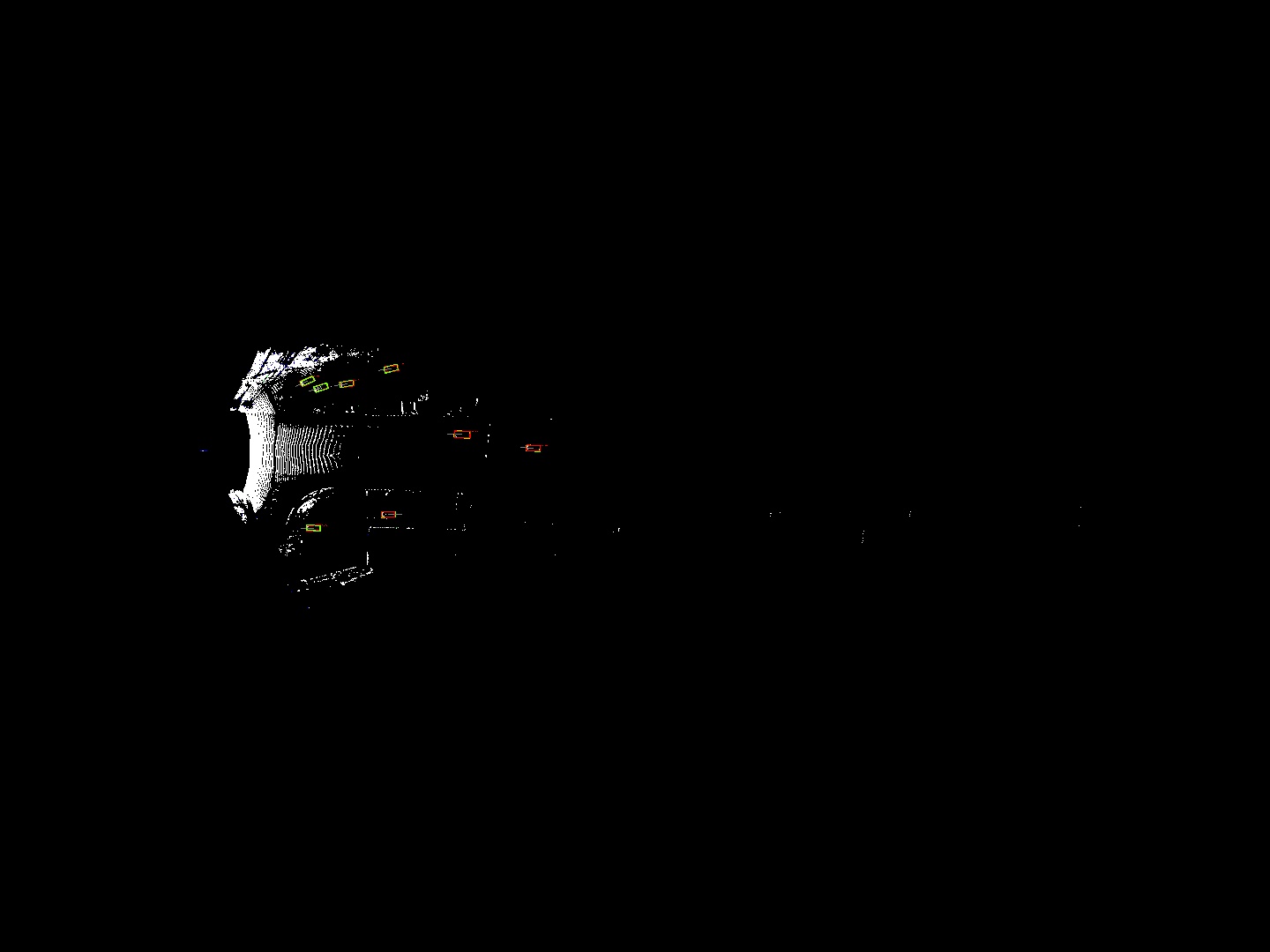}
    \includegraphics[trim=220 380 840 350, clip=true, width=0.327\linewidth]{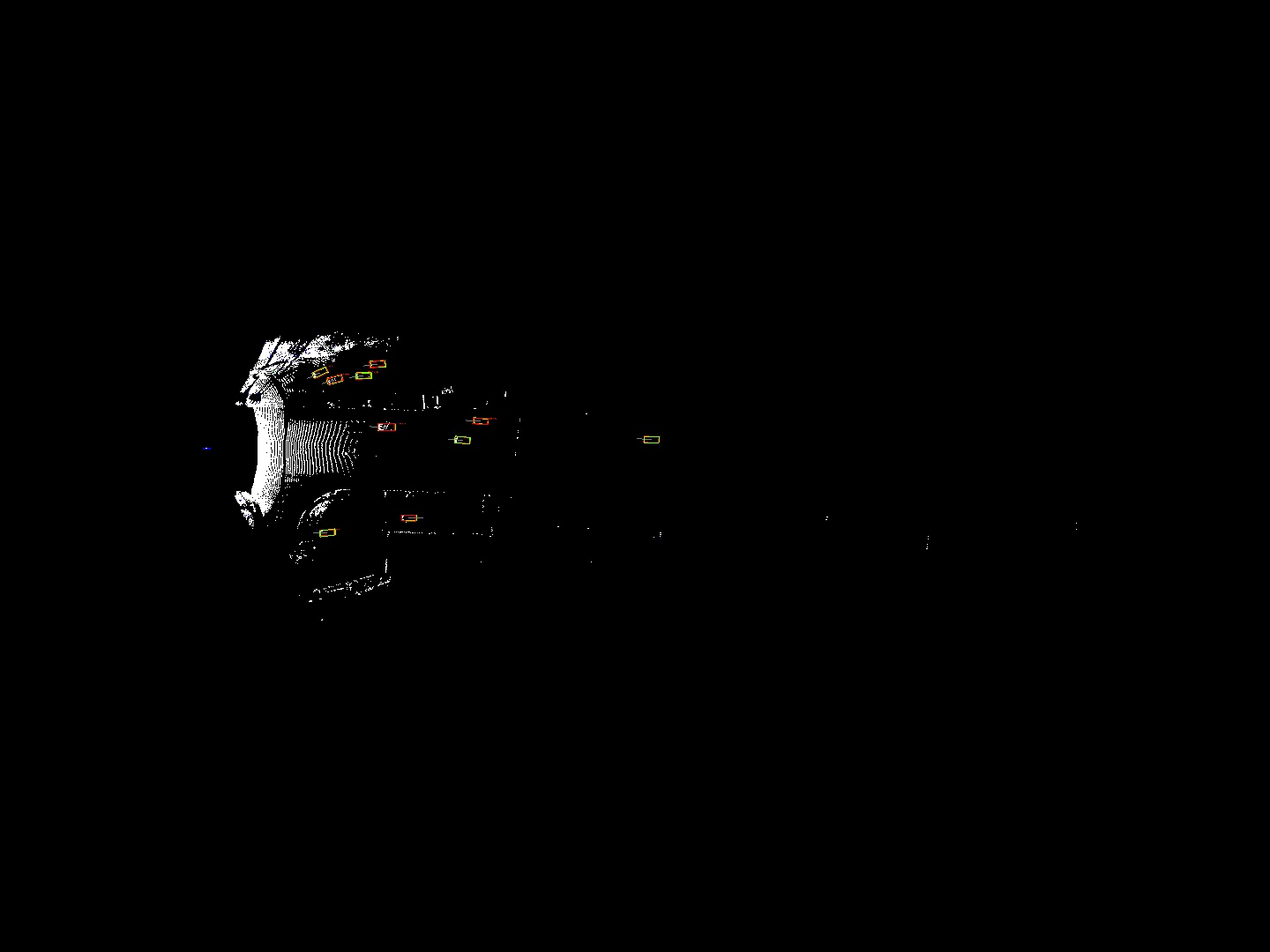}
    \includegraphics[trim=220 380 840 350, clip=true, width=0.327\linewidth]{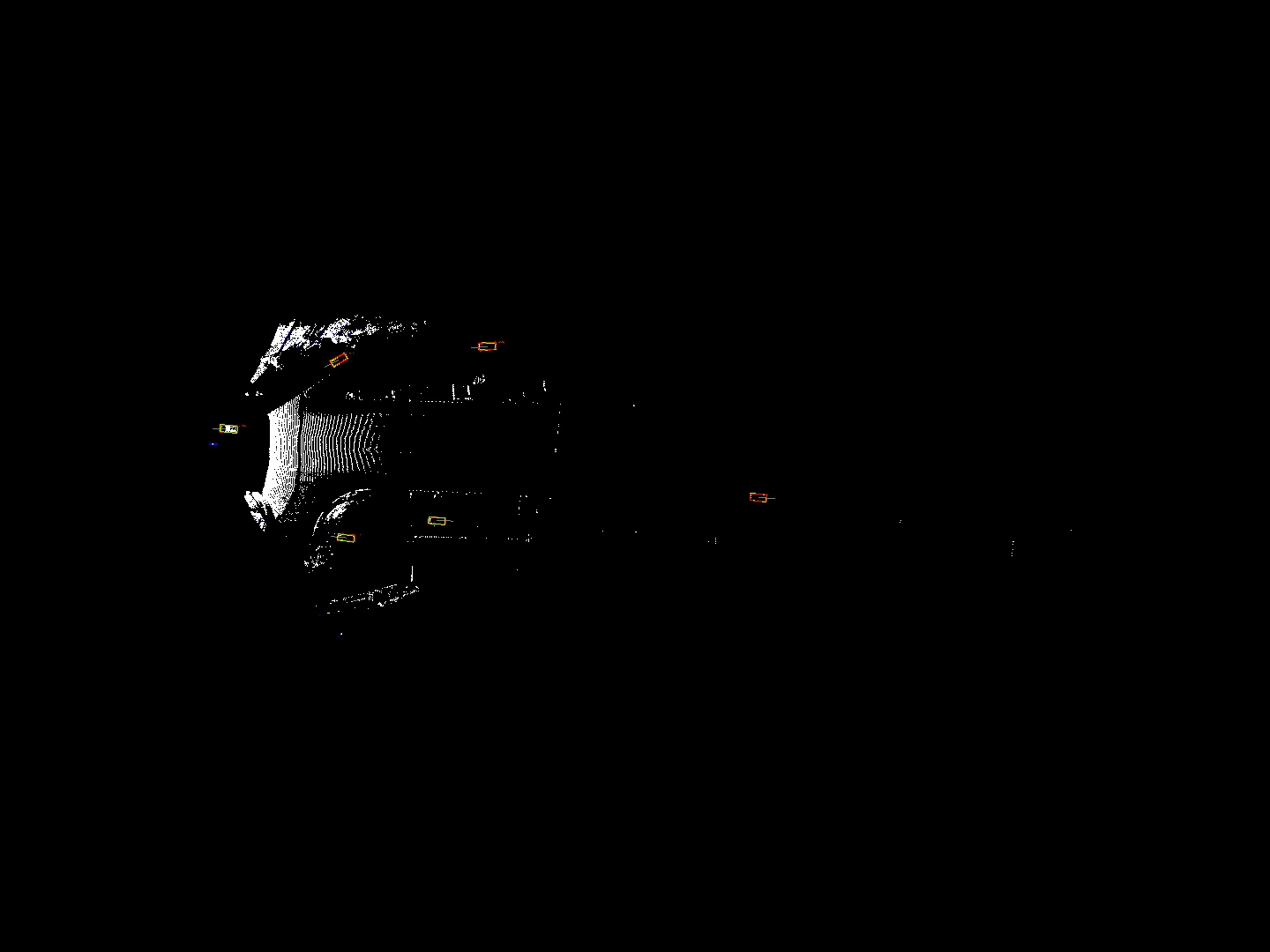} \\
    \includegraphics[trim=240 310 840 400, clip=true, width=0.327\linewidth]{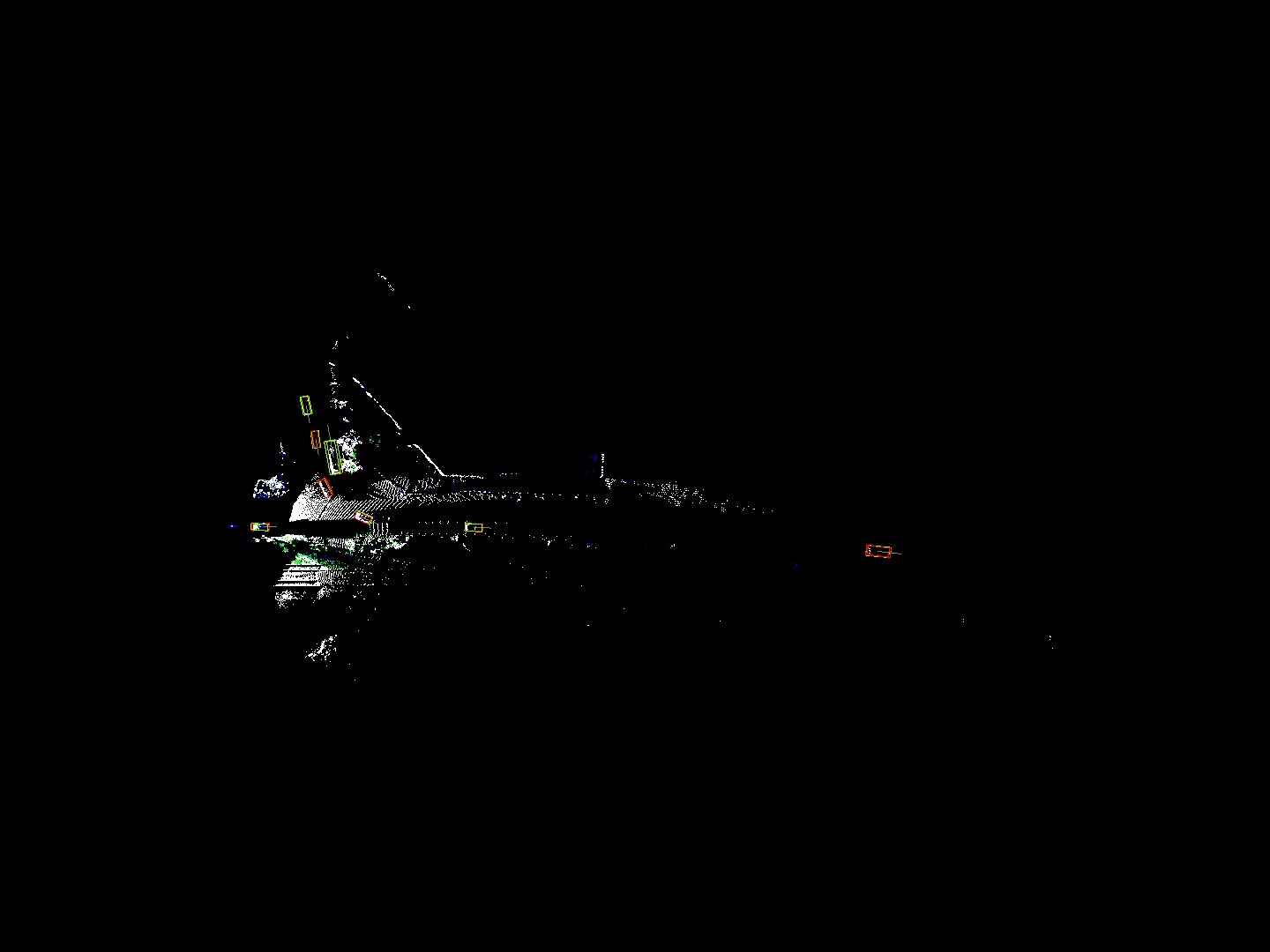}
    \includegraphics[trim=240 310 840 400, clip=true, width=0.327\linewidth]{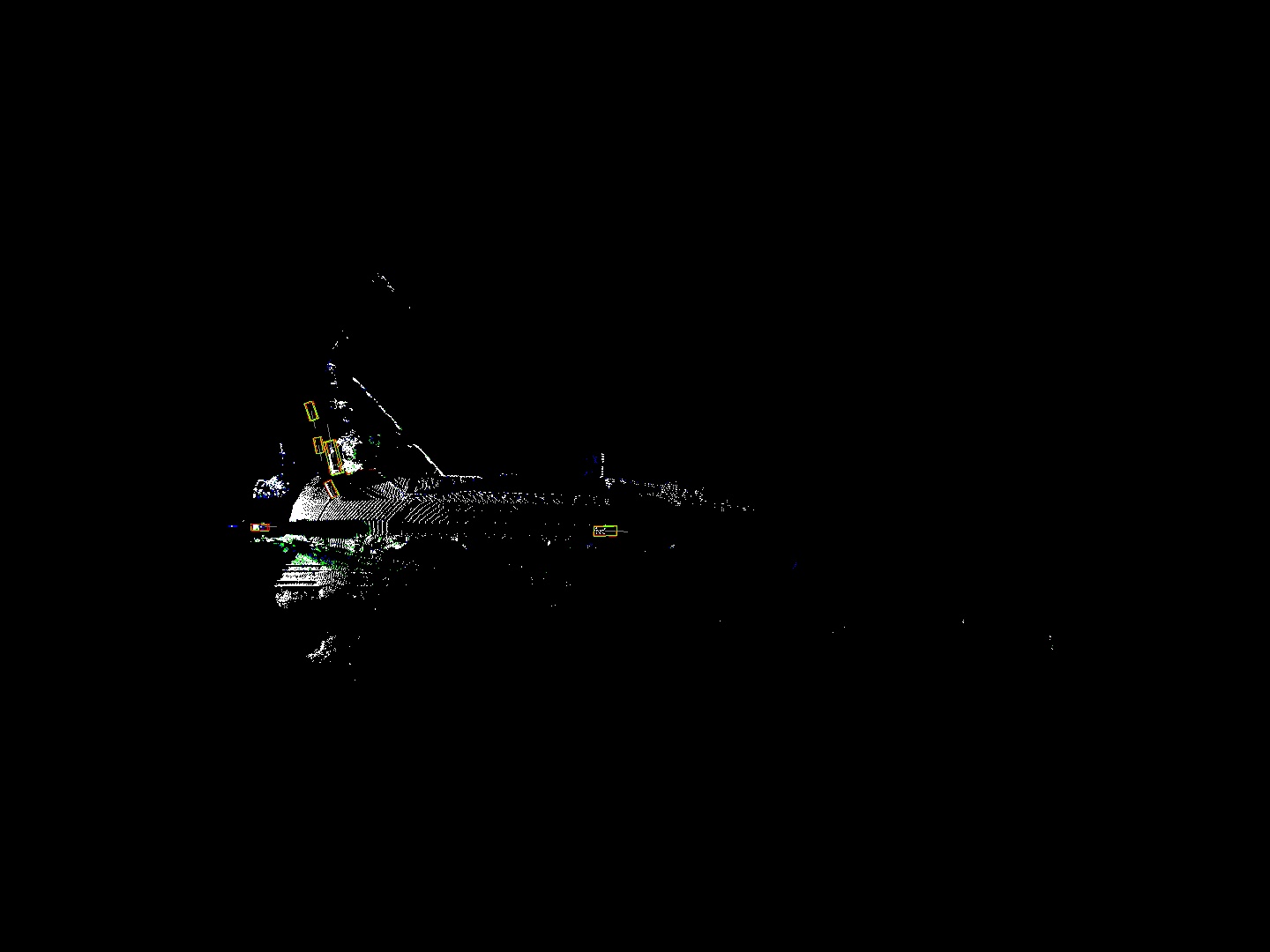}
    \includegraphics[trim=240 310 840 400, clip=true, width=0.327\linewidth]{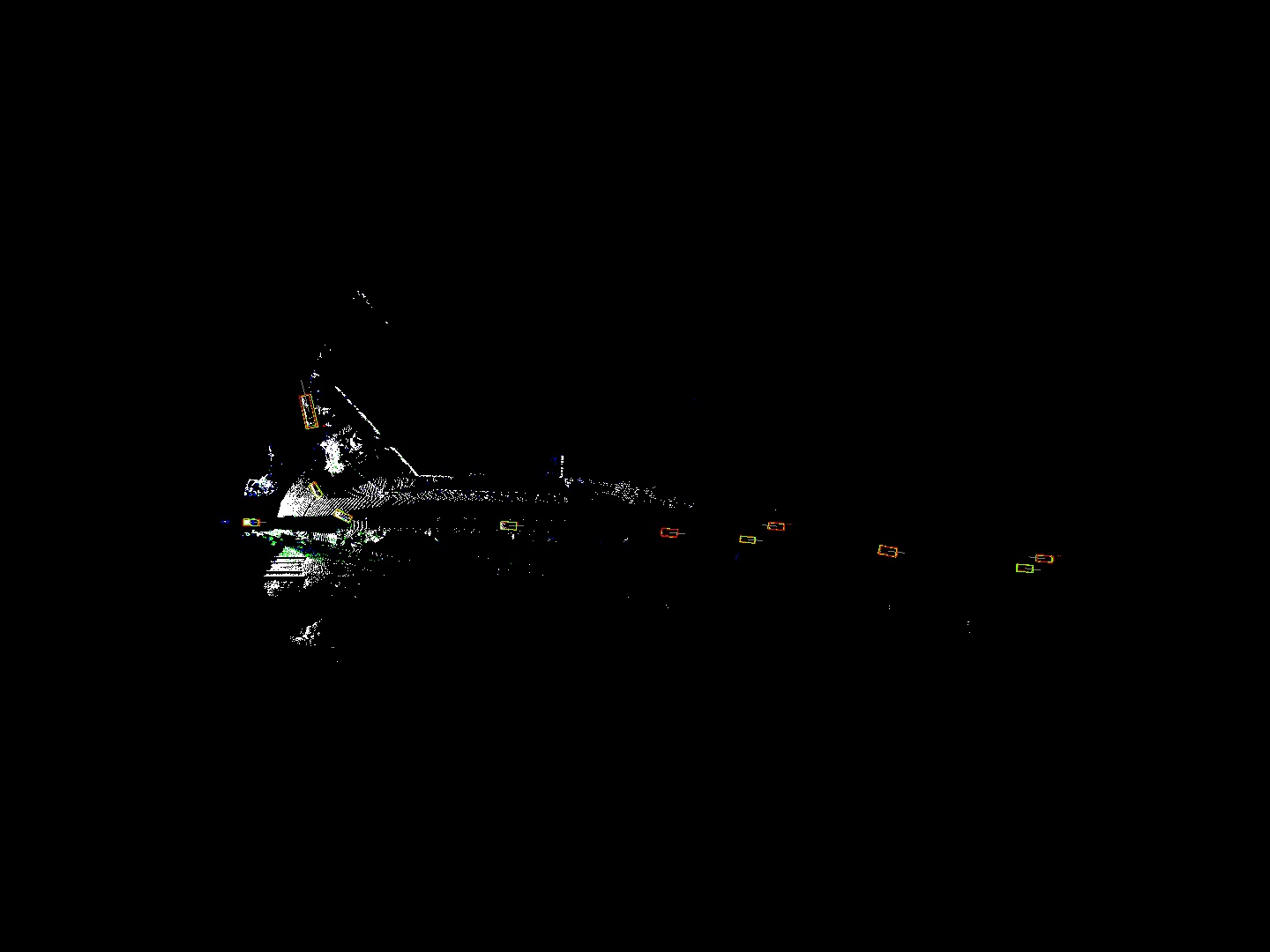} \\
    \vspace{-0.3cm}
    \caption{\textbf{Qualitative results of our MSLiD on real dataset.} The ground truth bounding boxes and predicted bounding boxes are labeled as red and green respectively. Note that \nth{1} echo points are shown as white, \nth{2} echo points are shown as green, and \nth{3} echo points are shown as blue.}
    \label{fig:qualitative}
    \vspace{-0.45cm}
\end{figure*}

\subsection{Multi-echo Point Cloud and Reflectance Signals}
\label{supp:mepc}

\vspace{1.5mm}\noindent\textbf{Generate Raw 3D Tensor Data.} Because the LiDAR sensor is essentially a time-of-flight measurement of photons, to simulate multi-signal LiDAR sensing measurements, we first simulate the photon measurements. Specifically, without considering the random false detection which happens occasionally in real LiDAR sensors, we formulate the number of photons $\mathbf{N}_{\mathbf{p}}$ received by a LiDAR detector in response to an illumination period of a signal light pulse by a temporal histogram:
\vspace{-0.1cm}
\begin{equation}
\mathbf{N}_{\mathbf{p}}[n] \sim \mathcal{P}(N_\mathsf{signal}[n]+ N_\mathsf{ambient}[n]) \label{eq:spad_eq},
\vspace{-0.1cm}
\end{equation} 
where $n$ is the $n$-th time interval along the temporal axis. Function $\mathcal{P}(\cdot)$ models a Poisson distribution, $N_\mathsf{signal}[n]$ is the number of detected signal photons at the time interval $n$, and $N_\mathsf{ambient}$ models the number of ambient photons.

Based on the Eq. \ref{eq:spad_eq}, the first step of our raw multi-echo LiDAR data generation is to generate a 3D tensor of photon counts $\mathbf{N}_{\mathbf{p}}[R_V, R_H, N]$ representing the number of photons detected by the sensor. Here $(R_V, R_H)$ is the vertical and horizontal resolution of the LiDAR (height and width of the `LiDAR Image') and $N$ represents the number of time intervals. 

To model the number of signal photons $N_\mathsf{signal}[n]$, we consider the surface reflectance, angle of incidence during reflection and radial attenuation. We model the relative photon number by assuming all LiDAR transmitters emit lasers with the same energy (same number of photons). Then, according to Lambert's cosine law, the reflected energy is proportional to $\cos(\theta)$ where $\theta$ is the incidence angle of the light with respect to the surface. This information is given by surface normal image $I_n^\prime$. We use a near infrared signal light, \emph{i.e}, the R channel of the RGB image $I_r^\prime$, to approximate the reflectance of the surface. Also, the radial falloff (attenuation) of light energy is proportional to the square of travel distance. We can directly take advantage of the accurate depth image $I_d^\prime$. Then, the number of signal photons is modeled as:
\vspace{-0.15cm}
\begin{equation}
\resizebox{0.905\hsize}{!}{
    $
    \mathbf{N}_{\mathsf{signal}}(h, w, n) \sim \left\{
        \begin{array}{ll}
              \text{Norm}\left(\mathsf{SBR} \times \frac{I_r^\prime(h, w)\cdot \cos\theta}{I_d^\prime(h, w)^2} \right) & \text{If  } n = n^* \\
              0 & \text{If  } n \neq n^* \\
        \end{array} 
    \right.
    $
}
\label{eq:spad_sim1}
\vspace{-0.1cm}
\end{equation} 
where the $\text{Norm}$ operation means to normalize over the whole image, (\emph{i.e.} divided by the average value in the entire image), $\mathsf{SBR}$ is the Signal-Background-Ratio used to control the relative strength between signal and background light, and $n^*$ is the time bin during which the signal light is reflected by the surface.

To model the number of ambient photons $N_\mathsf{ambient}[n]$, we simply takes the R-channel of the RGB images and normalize over the whole image,
\begin{equation}
\resizebox{0.9\hsize}{!}{
$N_\mathsf{ambient}(h, w, n) \sim \mathsf{Norm}\left(I_{r}[h, w]\right), \ \ \forall n \in [1, N]$
}
\label{eq:spad_sim2}
\vspace{-0.1cm}
\end{equation}
Then using Eq.~\ref{eq:spad_eq} together with Eq.~\ref{eq:spad_sim1} and Eq.~\ref{eq:spad_sim2}, we can simulate the 3D tensor of photon number $\mathbf{N}_{\mathbf{p}}[R_V, R_H, N]$.


Given the 3D tensor of photon numbers, the next step is to model the multi-echo mechanism. As explained in the main paper, multiple echoes happen because laser beams have a wider coverage of the 3D space instead of a perfect 2D line. In the 2D `LiDAR Image' space, this can be explained as -- Neighbor pixels overlap with each other.

We use a kernel function $G(\cdot)$ to simulate the spatial coverage, \emph{i.e.}, the number of photons in a given time bin will be a weighted sum of its spatial neighborhood bins (not temporal ones), with nearer neighbors contributing more:
\vspace{-0.15cm}
\begin{equation}
    \mathbf{N}^*_{\mathbf{p}}[h, w, n] = \mathop{\sum\sum}_{(h,w) \in \mathcal{N}(h, w)} G(k_h, k_w)\cdot \mathbf{N}_{\mathbf{p}}[h, w, n],
\label{eq:spad_ff}
\end{equation} 
where $\mathcal{N}$ is the neighborhood of a given position on the image plane, and $G(k_h, k_w)$ is the weight function over the distance between a given 2D position $(r_h, r_w)$ and its neighbor position $(k_h, k_w)$. Specifically, we use a Gaussian function for $G(\cdot)$. By controlling the parameters of the kernel function, we can control the spatial coverage of the lasers.

\vspace{1.5mm}\noindent\textbf{Generate Top-K Point Cloud Returns.} Because standard LiDAR-based perception systems \cite{sws20parta2,yml18second,lvc19pointpillars} take point cloud data as input (not raw photon count data $\mathbf{N}^*_{\mathbf{p}}[R_V, R_H, N]$), we take one step further to convert our raw multi-echo LiDAR data into point clouds so that they can be easily used in modern perception systems. Note that, with a large spatial coverage rate, each laser beam is able to cover a large 3D volume and is more likely to hit more than one target (object). This is represented as multiple strong peaks along the temporal histogram of a sensor beam. Specifically, when generating the top-K point cloud returns, we take the top-K maximum bins along the temporal axis for each sensor beam, \emph{i.e.}, we select the bin with the top-K number of photons that exceeds a threshold and obtain $\mathbf{N}^*_{\mathbf{p}}[R_V, R_H, K]$ as follows:
\vspace{-0.1cm}
\begin{equation}
\resizebox{0.9\hsize}{!}{
$\mathbf{N}^*_{\mathbf{p}}[R_V, R_H, K] = T\left(S\left(\mathbf{N}^*_{\mathbf{p}}[R_V, R_H, N]\right)[\text{:, :, :K}]\right)$,
}
\label{eq:spad_top}
\vspace{-0.1cm}
\end{equation} 
where $S(\cdot)$ is a sort function that descendingly sorts the number of photon counts along the temporal axis $N$. Then we only take the top-K channels in the temporal axis (\emph{i.e.}, the top-k maximum bins). Also, $T(\cdot)$ is a threshold function that masks out bins with number of photons less than a threshold, \emph{i.e.}, reject bins that receive noise instead of a light signal. Once we have obtained the $\mathbf{N}^*_{\mathbf{p}}[R_H, R_V, {K}]$, we can then transform it to $K$ point clouds as each valid bin (non-rejected) can be back-projected to a point in 3D space. As we use a threshold to mask out invalid points, the number of valid points will be fewer when $K$ is higher, \emph{i.e.}, the \nth{1} strongest point cloud has more points than the \nth{2} strongest point cloud and so on. 

After getting the multi-echo point cloud, we can simulate the reflectance of each point by normalizing the number of points inside the bins, because we assume that each LiDAR sensor transmitter emits same number of photons. The reflectance values of the multi-echo point cloud can be rearranged into the `LiDAR Image' space $\mathbf{I_{\reflectance}}[R_V, R_H, K]$ .

We summarize our multi-echo LiDAR simulation process in Algorithm \ref{alg} and in Figure \ref{fig:spad_lidar}. In terms of the implementation details, we use our all five camera viewpoints and create a $360^\circ$ FoV of the multi-echo LiDAR point cloud. We use $10240$ numbers of time bins to voxelize $1000$ meters of depth range in each view. When we sample the LiDAR sensor array in the polar coordinate, each view has a $45^\circ$ vertical FoV and $140^\circ$ horizontal FoV. For frontal view, we sample in $[-20^\circ, 25^\circ]$ vertical FoV with $0.2^\circ$ of resolution, and $[-70^\circ, 70^\circ]$ horizontal FoV with $0.1^\circ$ of resolution. To simulate a relatively large spatial coverage, we define a $[5, 5]$ patch in image coordinate centered around the projected position as its neighborhood. 

\section{Qualitative results of MSLiD}

In the Sec. 4.3 of the main paper, we give a quantitative analysis of our method on the real-world dataset. Here, we shown some qualitative results of our MSLiD in Fig~\ref{fig:qualitative}. Note that \nth{1} echo points are shown as white, \nth{2} echo points are shown as green, and \nth{3} echo points are shown as blue. Our method achieves good results on the real-world dataset in different traffic and scenarios.

\end{document}